\documentclass{article}
\usepackage{arxiv}

\usepackage[utf8]{inputenc} % allow utf-8 input
\usepackage[T1]{fontenc}    % use 8-bit T1 fonts
\usepackage{hyperref}       % hyperlinks
\usepackage{url}            % simple URL typesetting
\usepackage{booktabs}       % professional-quality tables
\usepackage{amsfonts}       % blackboard math symbols
\usepackage{nicefrac}       % compact symbols for 1/2, etc.
\usepackage{microtype}      % microtypography
\usepackage{graphicx}

\usepackage[table]{xcolor}
\usepackage{epstopdf}
\usepackage{algorithmic}
\usepackage{algorithm}
\usepackage[footnotesize]{caption}
\usepackage{subcaption}
\usepackage{bbold}
\usepackage{multirow}
\usepackage{tikz}
\usepackage{pgfplots}
\usepackage{xcolor}
\usepackage{makecell}
\usepackage{adjustbox}
\usepackage{longtable}
\usepackage{amsmath}
\usepackage{cleveref}
\usepackage[margin=1cm]{caption}
\usepackage{listings}

\DeclareSymbolFontAlphabet{\amsmathbb}{AMSb}

\usetikzlibrary{positioning}
\usepackage{tikz-cd}

\allowdisplaybreaks

\ifpdf
  \DeclareGraphicsExtensions{.eps,.pdf,.png,.jpg}
\else
  \DeclareGraphicsExtensions{.eps}
\fi

\newcommand{\norm}[1]{\left\lVert#1\right\rVert}
\DeclareMathOperator*{\minB}{min}

\DeclareMathOperator{\tr}{tr}

\newcolumntype{P}[1]{>{\centering\arraybackslash}p{#1}}
\usepackage{tabularx}
\newcolumntype{L}{>{\raggedright\arraybackslash}X}

\title{Manifold attack}

\author{
 Khanh-Hung Tran \\
  Institut LIST, CEA \\
  Université Paris-Saclay\\
  Gif-sur-Yvette, 91191 \\
  \texttt{khanh-hung.tran@cea.fr} \\
   \And
Fred Maurice Ngole Mboula \\
  Institut LIST, CEA \\
  Université Paris-Saclay\\
  Gif-sur-Yvette, 91191 \\
  \texttt{fred-maurice.ngole-mboula@cea.fr} \\
  \And
Jean-Luc Starck \\
  Astrophysics Department, CEA\\
  Université Paris-Saclay\\
  Gif-sur-Yvette, 91191 \\
  \texttt{jean-luc.starck@cea.fr} \\
}

\begin{document}
\maketitle
\begin{abstract}
Machine Learning in general and Deep Learning in particular has gained much interest in the recent decade and has shown significant performance improvements for many Computer Vision or Natural Language Processing tasks. In this paper, we introduce \textit{manifold attack}, a combination of manifold learning and adversarial learning, which aims at improving neural network models regularization. We show that applying \textit{manifold attack} provides a significant improvement for robustness to adversarial examples and it even enhances slightly the accuracy rate. Several implementations of our work can be found in \textcolor{blue}{\url{https://github.com/ktran1/Manifold-attack}}.
\end{abstract} 


\section{Introduction}

Deep Learning (DL) \cite{LeCun15DL} has been first introduced by Alexey Ivakhnenko (1967), then its derivative Convolutional Neural Networks \cite{Fukushima80} (1980) has been introduced by Fukushima. Over the years, it was improved and refined in by Yann LeCun \cite{Lecun98} (1998). Up to now, deep neural networks have yield groundbreaking performances at various classification tasks \cite{Krizhevsky12,simonyan2014deep,HeKaiming15,Szegedy14,Hinton12SpeechRe}. However, training deep neural networks involves different regularization techniques, which are primordial in general for two goals: generalization and adversarial robustness. On the one hand, regularization for generalization aims at improving the accuracy rate on the data that have not been used for training. In particular, this regularization is critical when the number of training samples is relatively small compared to the complexity of neural network model. On the other hand, regularization for adversarial robustness aims at improving the accuracy rate with respect to adversarial samples. %These adversarial samples can be generated from observed data or a learned manifold of observed data by adversarial transformations. 

\begin{figure}[h]
\centering
  \includegraphics[width=0.6\linewidth]{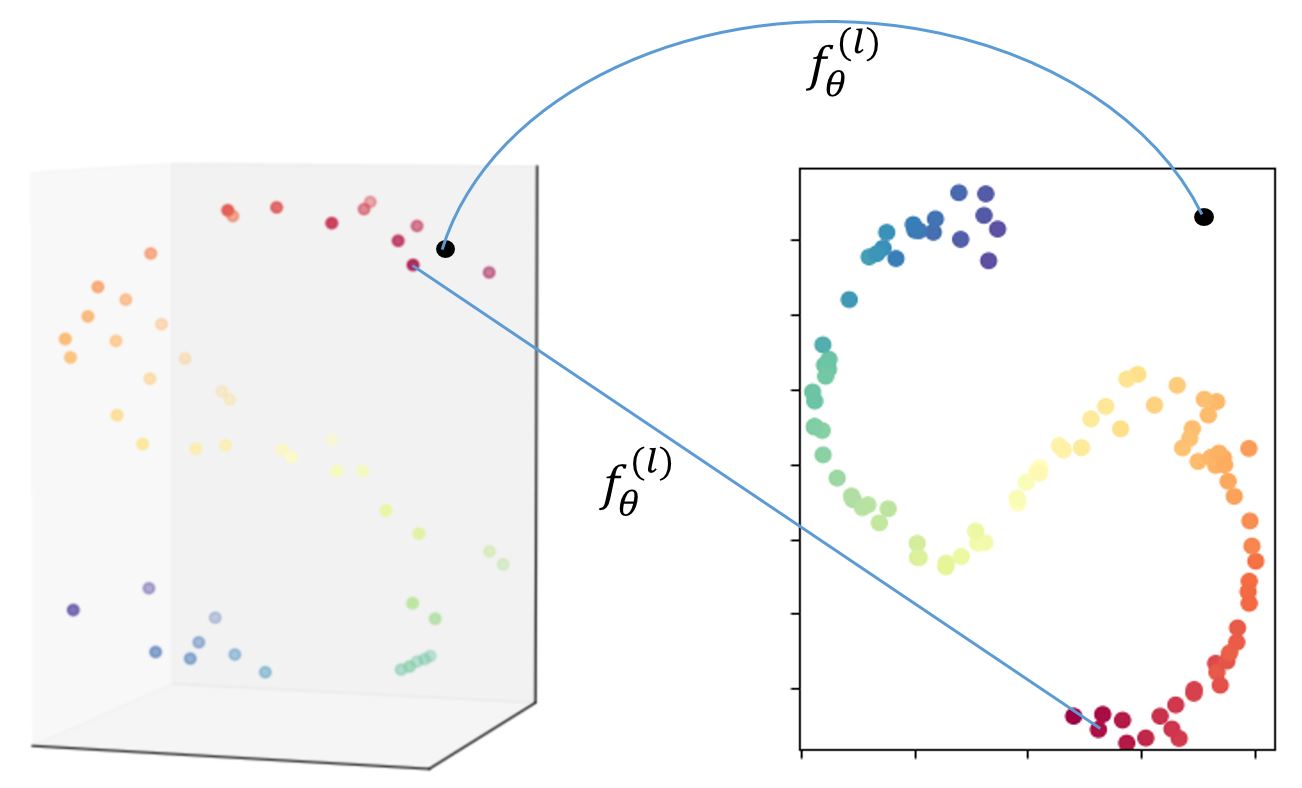}
  \caption{Preservation of geometric structure and a failure example. On the left, the original representation of S curve data (3 dimensions). On the right, an embedding of these data in dimension 2). The similarity between samples is indicated by different colors.}
  \label{fig:S_forme_data_preservation}
\end{figure}

In the last few years, a lot of research works in machine learning focus on getting high performance models in terms of adversarial robustness without losing in generalization. Our work focuses on the preservation of geometric structure (PGS) between the original representation of data and a latent representation produced by a neural network model (NNM). PGS is part of manifold learning techniques: we assume that the data of interest lies on a low dimensional piecewise smooth manifold with respect to which we want to compute original samples embeddings. In neural network models (NNMs), the output of the hidden layers are considered to be candidate low dimensional embeddings. For classification tasks, combining PGS as a regularization loss with supervised loss was shown to improve generalization ability by Weston \textit{et al.} in 2008 \cite{Weston08}. The concept of adversarial samples was later introduced by Goodfellow \textit{et al.} \cite{Goodfellow14explaining} in 2014, as NNMs architecture became deeper and hence more complex. Adversarial robustness in NNMs is strictly related to the PGS, since PGS for NNMs precisely aims at ensuring that the NNM gives similar outputs for close inputs. Figure \ref{fig:S_forme_data_preservation} shows an illustration of PGS including a failure example. %Each point (sample) in the original representation is mapped to its latent representation by the neural network $f_{\theta}^{(l)}$, where $(l)$ is the hidden layer's index. Without taking into account the black point, geometric structure in the original representation is perfectly preserved in the latent representation because all these points are used for learning $f_{\theta}^{(l)}$. Assume that we have the black point which is not used for training $f_{\theta}^{(l)}$. Potentially, this point and a red point can form a couple of antagonist points, which means they are similar in the original representation but their latent representations are very far apart. 
The black point can be considered an adversarial example for the dimension reduction model.

The novelty of this paper is to reinforce PGS by introducing synthetic samples as supplementary data. We make use of adversarial learning methodology to compute relevant synthetic samples. Then, these synthetic samples are added to the original real samples for training the model $f_{\theta}$ which is expected to improve adversarial robustness.     

\textbf{Contributions} :
\begin{itemize}
    \item We present the concept of manifold attack to reinforce PGS task.
    \item We show empirically that by applying manifold attack for a classification task, NNMs get better in both adversarial robustness and generalization. %However, by applying manifold attack \textit{with strong intensity}, there exists a slight trade-off between these two goals. 
\end{itemize}

The outline of the paper is as follows. In \cref{sec:Related works}, we present related works that aim to increase adversarial robustness of NNMs. In \cref{sec:Preservation of geometric structure}, we revisit some classical PGS methods. In \cref{sec:Manifold attack}, we present manifold attack. The numerical experiments are presented and discussed in \cref{sec:Applications of manifold attack}. Our conclusions and perspectives are presented in \cref{sec:Conclusions}.

\section{Related works}
\label{sec:Related works}
 
%We start this section by presenting the concept of off-manifold and on-manifold. Next, we present several strategies to reinforce adversarial robustness. 
In this section, we present several strategies to reinforce adversarial robustness, sorting them into two categories.

%Off-manifold means for examples that leave the manifold of data. For example, in MNIST dataset, an image with noise at almost all pixels, both background and object, is off-manifold because each sample in this dataset does not contain noise. On the contrary, on-manifold mean for examples that lie on the manifold of data and they share the same characteristics as observed samples. For example, an example generated by generative model can be considered as an on-manifold example. 

The first category is related to off-manifold adversarial examples i.e. adversarial examples that lies off the data manifold. In a supervised setting, these adversarial examples are used for training model as additional data\cite{Goodfellow14explaining}. In a semi-supervised setting, these adversarial examples are used to enhance the consistency between representations \cite{Miyato17VAT}. In plain words, a sample and its corresponding adversarial example must have a similar representation at the output of model. However, some works \cite{Su2019IsRobustnessCostAccuracy, Tsipras2019RobustnessAtOddAccuracy} state that adversarial robustness with respect to off-manifold adversarial examples is in conflict with generalization. In other words, improving adversarial robustness is detrimental to generalization and vice versa. 

In the second category, the adversarial robustness is reinforced with respect to on-manifold adversarial examples \cite{Song2018constructing,Stutz2019RobustnessAndGeneralization,Dia2019Semantics}. In all these works, a generative model such as VAE\cite{Kingma13VAE}, GAN\cite{Goodfellow14GAN} or VAE-GAN \cite{Larsen2016VAE_GAN} is firstly learnt, so that one can generate a sample on the data manifold from parameters in a latent space. %With the generator of generative model, a sampling in this latent representation corresponds an on-manifold example in the original representation. 
These works differ in the technique used to create adversarial noise in the latent space to produce on-manifold adversarial examples. Finally, these on-manifold adversarial examples are used as additional data for training model. Interestingly, by using on-manifold adversarial examples, \cite{Stutz2019RobustnessAndGeneralization} states that adversarial robustness and generalization can be jointly enhanced. We note that a hybrid approach has been proposed in \cite{Lin2020DualManifoldAdversarial} where both off-manifold and on-manifold adversarial examples are used as supplement data for training model.  

Our method falls into the second category since we want to enhance both the adversarial robustness and the generalization in both supervised and semi-supervised settings. However the adversarial examples are generated using a different paradigm.

\section{Preservation of geometric structure}
\label{sec:Preservation of geometric structure}

In most cases, a PGS process has two stages: extracting properties of original representation then computing an embedding which preserves these properties into a low dimensional space. Given a data set $\mathcal{X} = \{\mathbf{x}_1,..., \mathbf{x}_N \}$, $\mathbf{x}_i \in \mathbb{R}^n$ and its embeddings set $ \mathcal{A} = \{\mathbf{a}_1,...,\mathbf{a}_N \}, \mathbf{a}_i \in \mathbb{R}^p$, where $\mathbf{a}_i$ is the embedded representation of $\mathbf{x}_i$, we note:
\begin{equation}
\label{equa:general_preservation_loss}
    \mathcal{L}_e(\mathbf{a}_i,\mathcal{A}_i^c),
\end{equation}
the embedding loss, where $\mathcal{A}_i^c$ is the complement of $\{\mathbf{a}_i\}$ in $\mathcal{A}$. The objective function of PGS is then defined as:  
\begin{equation}
\label{equa:Embedding_Loss_total_function}
    \minB_{\mathcal{A}}\mathcal{L}_t = \minB_{\mathcal{A}}\sum_{i=1}^N\mathcal{L}_e(\mathbf{a}_i,\mathcal{A}_i^c)  
\end{equation}

We present some popular embedding losses hereafter:

- \textit{Multi-Dimensional scaling} (MDS) \cite{Kruskal78}: 
\begin{equation}
\label{equa:Multidimensional_scaling_loss}
    \mathcal{L}_e(\mathbf{a}_i,\mathcal{A}_i^c) =  \sum_{\mathbf{a}_j \in \mathcal{A}_i^c}(d_a(\mathbf{a}_i,\mathbf{a}_j) - d_x(\mathbf{x}_i,\mathbf{x}_j))^2,
\end{equation}
where $d_a()$ and $d_x() $ are measures of dissimilarity. By default, they are both Euclidean distances. This method aims at preserving pairwise distances from the original representation in the embedding space.   

- \textit{Laplacian eigenmaps} (LE) \cite{Belkin03}: 
\begin{equation}
\label{equa:Laplacian_Eigenmaps_loss}
\mathcal{L}_e(\mathbf{a}_i,\mathcal{A}_i^c) = \sum_{\mathbf{a}_j \in \mathcal{A}_i^c}d_x(\mathbf{x}_i,\mathbf{x}_j)d_a(\mathbf{a}_i,\mathbf{a}_j), 
\end{equation} 
where $d_x()$ is a measure of similarity measure (for instance $d_x(\mathbf{x}_i,\mathbf{x}_j) = \text{exp} \big( \frac{-\norm{\mathbf{x}_i-\mathbf{x}_j}_2^2}{2\sigma_i^2} \big)$) and $d_a()$ is a measure of dissimilarity (for instance $ d_a(\mathbf{a}_i,\mathbf{a}_j) = \norm{\mathbf{a}_i - \mathbf{a}_j}_2^2$). This method learns manifold structure by emphasizing the preservation of local distances. In order to further reduce effect of large distances, in some papers, $d_x(\mathbf{x}_i,\mathbf{x}_j)$ is set directly to zero if $\mathbf{x}_j$ is not in the $k$ nearest neighbors of $\mathbf{x}_i$ or vice versa if $\mathbf{x}_i$ is not in the $k$ nearest neighbors of $\mathbf{x}_j$. Alternatively, one can set $d_x(\mathbf{x}_i,\mathbf{x}_j) = 0$ if $\norm{\mathbf{x}_i-\mathbf{x}_j}_2^2 > \kappa$. Let $\mathbf{W}$ be the matrix defined as $\mathbf{W}_{ij} = d_x(\mathbf{x}_i,\mathbf{x}_j)$ hence $\mathbf{W}$ is a symmetric matrix if $d_x()$ is symmetric. We can represent the objective function of Laplacian eigenmaps method using the matrices: 
\begin{equation}
\mathcal{L}_t = \sum_{i=1}^N\mathcal{L}_e(\mathbf{a}_i,\mathcal{A}_i^c)  = \sum_{i=1}^{N}\sum_{\substack{j=1 \\ j\neq i}}^{N} \mathbf{W}_{ij}\norm{\mathbf{a}_i - \mathbf{a}_j}_2^2 =2\tr(\mathbf{A}\mathbf{L}\mathbf{A}^\top),
\end{equation}
where $\mathbf{L} = \mathbf{D} - \mathbf{W}$, $\mathbf{D}_{ii} = \sum_{j=1}^{N} \mathbf{W}_{ij}$ ($\mathbf{D}$ being a diagonal matrix). $\mathbf{L}$ a graph Laplacian matrix because it is symmetric, the sum of each row equals to 1 and its elements are negatives except for the diagonal elements. 

- \textit{Locally Linear Embedding} (LLE) \cite{Roweis00}:
\begin{equation}
\label{equa:LLE_loss}
    \mathcal{L}_e(\mathbf{a}_i,\mathcal{A}_i^c) = \norm{\mathbf{a}_i - \sum_{\mathbf{a}_j \in \mathcal{A}_i^c } \lambda_{ij} \mathbf{a}_{j}}_2^2,  
\end{equation}
where $\lambda_{ij}$ are determined by solving the following problem:
\begin{equation}
\label{equa:LLE_lamda_opti}
\begin{split}
&\minB_{\lambda_{ij}} \norm{\mathbf{x}_i - \sum\limits_{j} \lambda_{ij} \mathbf{x}_{j}}_2^2,\\
\text{subject to: }&\begin{cases} 
\sum\limits_{j} \lambda_{ij} = 1, \text{ if } \mathbf{x}_j \in \text{knn}(\mathbf{x}_i),\\
 \lambda_{ij} = 0 \text{ if not.} \\
\end{cases} \\ 
\end{split}
\end{equation}
where knn($\mathbf{x}_i$) denotes a set containing indices of the $k$ nearest neighbors samples (in Euclidean distance) of the sample $\mathbf{x}_i$. Assuming that the observed data $\mathcal{X}$ is sampled from a smooth manifold and provided that the sampling is dense enough, one can assume that the sample lies locally on linear patches. Thus, LLE first computes the barycentric coordinate for each sample \textit{w.r.t.} its nearest neighbors. These barycentric coordinates characterize the local geometry of the underlying manifold. Then, LLE computes a low dimensional representation (embedded) which is compatible with these local barycentric coordinates. Introducing $\mathbf{V} \in \mathbb{R}^{N \times N}$ a matrix representation form for $\lambda$ as: $\mathbf{V}[j,i] = \lambda_{ij} $ then $\mathcal{L}_t = \sum_{i=1}^N\mathcal{L}_e(\mathbf{a}_i,\mathcal{A}_i^c) $ can be rewritten as $\mathcal{L}_t =  \norm{\textbf{A} - \textbf{A}\textbf{V}}_F^2 =   \tr(\mathbf{A}\mathbf{L}\mathbf{A}^\top)$, where $\mathbf{L} = \mathbf{I}_N-\mathbf{V}-\mathbf{V}^\top+\mathbf{V}^\top\mathbf{V}$. Thus, the loss $\mathcal{L}_t$ can be interpreted as Laplacian eigenmaps loss, based on an implicit metric $d_x$ for measuring distance between two samples.

% - \textit{Laplacian Learning} (LL) \cite{Dong16}:
% \begin{equation}
% \mathcal{L}_t = \tr(\mathbf{A}\mathbf{L}\mathbf{A}^\top),
% \end{equation}
% where $\mathbf{L}$ is learnt by solving the following problem:
% \begin{equation}
% \label{equa:Laplacian_Learning}
% \begin{aligned}
% & \minB_{\mathbf{L} \in \mathbb{R}^{N \times N}} & & \tr{(\mathbf{X}\mathbf{L}\mathbf{X}^{\top})}+\theta\norm{\mathbf{L}}_F^2 \\
% & \text{subject to}
% & &  \tr(\mathbf{L}) = N, \\
% & & & \mathbf{L}_{ij} = \mathbf{L}_{ji} < 0 (i \neq j), \\
% & & & \sum_j \mathbf{L}_{ij} = 0.
% \end{aligned}
% \end{equation}

% The constraint $\tr(\mathbf{L}) = N$ is set in order to control the energy of $\mathbf{L}$. Instead of creating $\mathbf{L}$ using a predefined generic metric as in LE method, in LL, the Laplacian matrix is learnt directly from the data. To optimize problem (\ref{equa:Laplacian_Learning}), we can use interior point method \cite{Boyd04Convex} or ADMM \cite{Boyd11ADMM}).

- \textit{Contrastive loss }:
\begin{equation*}
\label{equa:Contrastive_loss}
    \mathcal{L}_e(\mathbf{a}_i,\mathcal{A}_i^c) = \sum_{\mathbf{a}_j \in \mathcal{A}_i^c} \Big( d_x(\mathbf{x}_i,\mathbf{x}_j) d_a(\mathbf{a}_i,\mathbf{a}_j) + (1-d_x(\mathbf{x}_i,\mathbf{x}_j)) \text{ max}(0,\tau - d_a(\mathbf{a}_i,\mathbf{a}_j)) \Big),   
\end{equation*}
where $d_x(\mathbf{x}_i,\mathbf{x}_j)$ is a discrete similarity metric which is equal to 1 if $\mathbf{x}_j$ is in the neighborhood of $\mathbf{x}_j$ and 0 otherwise. $d_a()$ is a measure of dissimilarity. %by default $d_a(\mathbf{a}_i,\mathbf{a}_j) = \norm{\mathbf{a}_i - \mathbf{a}_j}_2^2$. 

- \textit{Stochastic Neighbor Embedding} (SNE) \cite{Hinton03SNE}:
\begin{equation*}
\label{equa:SNE_loss}
    \mathcal{L}_e(\mathbf{a}_i,\mathcal{A}_i^c) = \sum_{\mathbf{a}_j \in \mathcal{A}_i^c} P_{ij} \text{log} \frac{P_{ij}}{Q_{ij}},  
\end{equation*}
where $P_{ij} = \frac{d_x(\mathbf{x}_i,\mathbf{x}_j)}{\sum_{k\neq i} d_x(\mathbf{x}_i,\mathbf{x}_k)}$ and $Q_{ij } = \frac{d_a(\mathbf{a}_i,\mathbf{a}_j)}{\sum_{k\neq i} d_a(\mathbf{a}_i,\mathbf{a}_k)}$, $d_x$ and $d_a$ are both similarity metric. The objective of this method is to preserve the similarity between two distributions of pairwise distances, one in original representation and the other in embedded representation, by Kullback–Leibler (KL) divergence.    

Traditionally, PGS finds its applications in dimensionality reduction or data visualization, which refers to the techniques used to help the analyst see the underlying structure of data and explore it. For instance, \cite{Maaten08T_SNE} proposes a variant of SNE, which has been used in a wide range of fields. Classical nonlinear PGS methods do not require a mapping model, which is a function $g()$ with trainable parameters that maps a sample $\textbf{x}$ to its embedded representation $\mathbf{a}$ as $\mathbf{a} = g(\mathbf{x})$. $\mathbf{a}$ is directly used as the optimization variable. 

Finally, it is worth noting that for several PGS methods such as LE and LLE, some supplement constraints are required to avoid a trivial solution (for instance with all embedded points are collapsed into only one point). Usually, mean and co-variance constrains are applied: 
\begin{equation}
\label{equa:Manifold_learning_constraints}
\begin{gathered}
    \mathrm{m}(\mathbf{A}) = [\mu(\mathbf{A}[1,:]),..,\mu(\mathbf{A}[p,:])]^\top = \mathbb{0} \\
    \text{Cov}(\mathbf{A},\mathbf{A}) = \big(\mathbf{A}-\mathrm{M}(\mathbf{A})\big)\big(\mathbf{A}-\mathrm{M}(\mathbf{A})\big)^\top = \mathbf{I}_p \\
\end{gathered}
\end{equation}

\section{Manifold attack}
\label{sec:Manifold attack}

We introduce a new strategy called ``manifold attack'' to reinforce PGS methods with mapping model, by combining with adversarial learning. In the following $g()$ denotes a differentiable function that maps a sample $\mathbf{x} \in \mathbb{R}^n$ to its embedded representation $\mathbf{a} \in \mathbb{R}^p$.

\subsection{Individual attack point versus data points}

We define a \textit{virtual point} is a synthetic sample generated in such a way to be likely on the observed samples underlying manifold. An \textit{anchor point} is a sample used for generating a virtual point. 
An \textit{attack point} is a virtual point that maximises locally a chosen measure of model distortion. For example, attack point can be a sample perturbed with an adversarial noise.

We use the same notation as in section \ref{sec:Preservation of geometric structure}. Given a dataset $\mathcal{X} = \{\mathbf{x}_1,..., \mathbf{x}_N \}$ and the corresponding embedded set $ \mathcal{A} = \{\mathbf{a}_1,...,\mathbf{a}_N \}$,  $\mathcal{L}_e(\mathbf{a}_i,\mathcal{A}_i^c)$ denotes the embedding loss, $\mathcal{A}_i^c$ being the complement of $\{\mathbf{a}_i\}$ in $\mathcal{A}$. 
We consider the objective function of PGS  defined as $\mathcal{L}_t = \sum_{i=1}^N\mathcal{L}_e(\mathbf{a}_i,\mathcal{A}_i^c)$. Let's consider $p$ anchor points $\mathbf{z}_1, \mathbf{z}_2,..., \mathbf{z}_p \in \mathbb{R}^{n} $, a virtual point $\tilde{\mathbf{x}}\in \mathbb{R}^{n}$ is defined as: 
\begin{equation}
\label{equa:attack_point_with_anchors}
\begin{split}
&\tilde{\mathbf{x}} = \gamma_1\mathbf{z}_{1} + \gamma_2\mathbf{z}_{2} + ... + \gamma_p\mathbf{z}_{p}, \\
\text{ subject to: } &\gamma_1, \gamma_2,.., \gamma_p \geq 0, \\
& \gamma_1+ \gamma_2+...+\gamma_p = 1.
\end{split}
\end{equation}

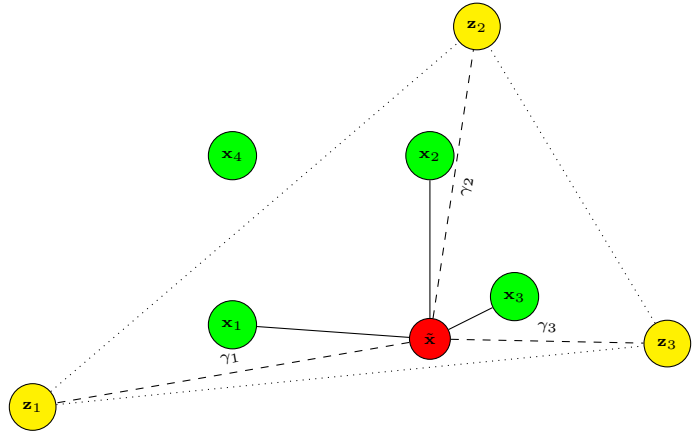
\begin{figure}[h]
\begin{center}
\begin{tikzpicture}[scale = 0.75]
\tikzstyle{every node}=[font=\tiny]

\tikzstyle{green dot}=[fill=green, draw=black, shape=circle ]
\tikzstyle{red dot}=[fill=red, draw=black, shape=circle ]
\tikzstyle{yellow dot}=[fill=yellow, draw=black, shape=circle ]

\def \ax{-3.0}
\def \ay{-1.5}
\def \bx{0.5} 
\def \by{1.5}
\def \cx{2.}
\def \cy{-1.0}
\def \s{2.25}

	\node [style=green dot] (0) at (\ax , \ay) {$\mathbf{x}_1$};
	\node [style=green dot] (1) at (\bx, \by) {$\mathbf{x}_2$};
	\node [style=green dot] (2) at (\cx, \cy) {$\mathbf{x}_3$};
	
	\node [style=green dot] (3) at (-3, 1.5) {$\mathbf{x}_4$};

	\node [style=red dot] (4) at (0.5, -1.75) {$\tilde{\mathbf{x}}$};
	
	\node [style=yellow dot] (5) at (2.0*\s*\ax/3.0 + \ax/3.0 + \bx/3.0 - \s*\bx/3.0 + \cx/3.0 - \s*\cx/3.0, 2.0*\s*\ay/3.0 + \ay/3.0 + \by/3.0 - \s*\by/3.0 + \cy/3.0 - \s*\cy/3.0) {$\mathbf{z}_1$};
	\node [style=yellow dot] (6) at (2.0*\s*\bx/3.0 + \bx/3.0 + \ax/3.0 - \s*\ax/3.0 + \cx/3.0 - \s*\cx/3.0, 2.0*\s*\by/3.0 + \by/3.0 + \ay/3.0 - \s*\ay/3.0 + \cy/3.0 - \s*\cy/3.0) {$\mathbf{z}_2$};
	\node [style=yellow dot] (7) at (2.0*\s*\cx/3.0 + \cx/3.0 + \bx/3.0 - \s*\bx/3.0 + \ax/3.0 - \s*\ax/3.0, 2.0*\s*\cy/3.0 + \cy/3.0 + \by/3.0 - \s*\by/3.0 + \ay/3.0 - \s*\ay/3.0) {$\mathbf{z}_3$};
	
    \draw [dotted] (5) to (6);
	\draw [dotted] (5) to (7);
	\draw [dotted] (7) to (6);

	\path[every node/.style={sloped,anchor=south,auto=false,font=\tiny}]
	
	(4) edge  (2)    
	(4) edge  (1)    
	(4) edge  (0)
% 	(4) edge  (3)
	
	(4) edge[dashed]  node [midway, above] {$\gamma_1$} (5)
	(4) edge[dashed]  node [midway, below] {$\gamma_2$} (6)
	(4) edge[dashed]  node [midway, above] {$\gamma_3$} (7);
	
\end{tikzpicture}
\end{center}
\caption{Virtual point and anchor points illustration. The three anchor points ($\mathbf{z}_i, i \in \{1,2,3\}$) are computed as $\mathbf{z}_i = \mu([\mathbf{x}_1,\mathbf{x}_2,\mathbf{x}_3]) + s(\mathbf{x}_i-\mu([\mathbf{x}_1,\mathbf{x}_2,\mathbf{x}_3])$, where $\mu([\mathbf{x}_1,\mathbf{x}_2,\mathbf{x}_3]) = \frac{\mathbf{x}_1 +\mathbf{x}_2 + \mathbf{x}_3}{3} $. The dotted lines represent the zone defined by anchor points $\mathbf{z}_i$, within which the virtual point $\tilde{\mathbf{x}}$ necessarily lies. The parameter $s$ determines whether the anchor points lie strictly within $[\mathbf{x}_1,\mathbf{x}_2,\mathbf{x}_3]$ convex hull $(1 > s > 0)$ or are strictly outside $(s > 1)$. The dashed lines represent the coordinates $\gamma$ of $\tilde{\mathbf{x}}$. The solid lines illustrates the relatedness of $\tilde{\mathbf{x}}$ to the data points that the embedding will try to preserve minimizing $\mathcal{L}_e(g(\tilde{\mathbf{x}}), \{g(\mathbf{x}_1),g(\mathbf{x}_2),g(\mathbf{x}_3)\})$ in this case. }
\label{fig:individual manifold attack}
\end{figure}

The anchor points $\mathbf{z}_i$ define a region or feasible zone, in which a virtual point $\tilde{\mathbf{x}}$ must be located and $\gamma = [\gamma_1, \gamma_2,.., \gamma_p]$ is the coordinate of $\tilde{\mathbf{x}}$. Anchor points are sampled from the dataset $\mathcal{X}$ with different strategies, which are defined according to predefined rules (see \cref{subsec:Settings of anchor points} for several examples). Figure \ref{fig:individual manifold attack} shows an example of anchor points setting and relations between points. For the embedding $\mathbf{a}_i$ of observed sample $\mathbf{x}_i$, $\mathbf{a}_i = g(\mathbf{x}_i)$, the embedding loss is defined as: 
\begin{equation}
    \mathcal{L}_e(\mathbf{a}_i,\mathcal{A}_i^c) = \mathcal{L}_e\big(g(\mathbf{x}_i),\{g(\mathbf{x}_1),..,g(\mathbf{x}_N)\} \backslash \{g(\mathbf{x}_i)\}\big). 
\end{equation}
Similarly, for the embedding $\tilde{\mathbf{a}}$ of virtual point $\tilde{\mathbf{x}}$, $\tilde{\mathbf{a}} = g(\tilde{\mathbf{x}})$, the embedding loss is defined as: 
\begin{equation}
\mathcal{L}_e(\tilde{\mathbf{a}},\mathcal{A}) = \mathcal{L}_e\big(g(\tilde{\mathbf{x}}), \{g(\mathbf{x}_1),..,g(\mathbf{x}_N)\}\big).
\end{equation}

The \cref{algorithm:individual_manifold_attack} describes the computation of an attack point. It consists in finding the local coordinates $\gamma$ that maximizes the embedding loss $\mathcal{L}_e(\tilde{\mathbf{a}},\mathcal{A})$ for the current state of model $g()$. %Finding attack point $\tilde{\mathbf{x}}$ is equivalent to finding $\gamma$. 
%The later is first initialized, then updated by adding gradient $\nabla_\gamma\mathcal{L}_e$, in order to give a local maximum of $\mathcal{L}_e$.
Hence, $\gamma$ is estimated though a projected gradient ascent.

\begin{algorithm}
\caption{Individual manifold attack}
\label{algorithm:individual_manifold_attack}
\begin{algorithmic}
\REQUIRE Anchor points $\{\mathbf{z}_1,..,\mathbf{z}_p\}$, data points $\{ \mathbf{x}_1,..,\mathbf{x}_N\}$, embedding loss $\mathcal{L}_e()$, model $g()$, $\xi,n\_iters$.
\STATE \textbf{initialize}: $\gamma \in \mathbb{R}^p, \gamma = [\gamma_1,.., \gamma_p]$ for constraints in \cref{equa:attack_point_with_anchors}
\STATE $\tilde{\mathbf{x}} = \gamma_1\mathbf{z}_{1} + \gamma_2\mathbf{z}_{2} + ... + \gamma_p\mathbf{z}_{p}$
\FOR{$i = 1$ \textbf{to} $n\_iters$}
\STATE $L =  \mathcal{L}_e \big(g(\tilde{\mathbf{x}}), \{g(\mathbf{x}_1),..,g(\mathbf{x}_N)\} \big) $
\STATE  $ \gamma \leftarrow \gamma + \xi \nabla_\gamma L(\tilde{\mathbf{x}})  $
\STATE  $\gamma \leftarrow \Pi_{ps}(\gamma)$
\STATE $\tilde{\mathbf{x}} = \gamma_1\mathbf{z}_{1} + \gamma_2\mathbf{z}_{2} + ... + \gamma_p\mathbf{z}_{p}$
\ENDFOR
\STATE \textbf{Output: $\tilde{\mathbf{x}}$} 
\end{algorithmic}
\end{algorithm}

In order to guarantee the constrains in equation (\ref{equa:attack_point_with_anchors}), we use the projector $\Pi_{ps}$ defined by the problem:

\begin{equation}
\label{equa:prox_operation_for_positif_and_constraint}
\begin{split}
&\min_{\gamma \in \mathbb{R}^p} \frac{1}{2} \norm{\kappa-\gamma}_2^2,\\
\text{ subject to: } &\gamma_1, \gamma_2,.., \gamma_p \geq 0, \\
& \gamma_1+ \gamma_2+...+\gamma_p = c, (c > 0). \\
\end{split}
\end{equation}

This convex problem with constraints can be solved quickly by a simple sequential projection that alternates between sum constraint and positive constraint (\cref{algorithm:Projection_for_positive_and_sum}). The demonstration can be inspired by Lagrange multiplier method. %This algorithm can be extended for executing in parallel multiple attack points at the same time ($\gamma \in \mathbb{R}^{M \times p}$, where $M$ is the number of attack points). 
A simple demonstration can be found in \nameref{sec:Appendix}.

\begin{algorithm}
\caption{Projection for positive and sum constraint $\Pi_{ps}$}
\label{algorithm:Projection_for_positive_and_sum}
\begin{algorithmic}
\REQUIRE $\kappa \in \mathbb{R}^p$, $c = 1$ (by default).
\STATE $\delta = (c - \sum_{i=1}^{p} \kappa_i)/p$
\STATE $\gamma_i \leftarrow \gamma_i +  \delta, \forall i = 1,..,p$
\WHILE{$\exists i \in \{1,..,p\}: \gamma_i <0 $}
\STATE $\mathcal{P} = \{i | \gamma_i > 0 \}$ and $ \mathcal{N} = \{i | \gamma_i <0 \}$
\STATE  $\gamma_i \leftarrow 0, \forall i \in \mathcal{N}$ 
\STATE  $\delta = (c -\sum_{i \in \mathcal{P}} \gamma_i)/|\mathcal{P}| $
\STATE  $\gamma_i \leftarrow \gamma_i + \delta, \forall i \in \mathcal{P}$
\ENDWHILE
\STATE \textbf{Output: $ \gamma = [\gamma_1,\gamma_2,..,\gamma_p]$} 
\end{algorithmic}
\end{algorithm}

By default, each set of anchor points has one attack point. However, we can generate more than one attack point for the same set of anchor points by using different initializations of $\gamma$, so as to find different local maxima. The double embedding loss, on the observed samples and on the attack points, is expected to enforce the the model $g()$ smoothness over the underlying manifold, including in low samples density areas. The general optimization scheme goes as follows: we optimize alternately between attack stages and model update stages until convergence. In attack stage, we optimize the attack points through $\gamma$ while fixing the model $g()$ and in the model update stage, we optimize the model $g()$ while fixing attack points.

% We create and find attack points that distort the embedded manifold of data, \textit{i.e.} we maximize the loss function $\mathcal{L}_t$  while fixing the current model $g()$, then we update the model to minimize this loss function while fixing the attack points

\subsection{Attack points as data augmentation}
In \cref{algorithm:individual_manifold_attack}, an attack points only interacts with observed samples. In the general manifold attack (algorithm \ref{algorithm:manifold_attack}), attack points and observed samples are undifferentiated in the model update stage. This way, hence generating attack points can be considered a data augmentation technique. We denote $\mathcal{B}$ as a set that contains all embedded points (both attack points and observed samples). $\mathcal{B}_s$ is a random subset of $\mathcal{B}$, used batch for batch optimization. In each step, only attack points from the current batch are used to distort the manifold by maximizing the batch loss $L$. %The effect between attack points and data points can be controlled via the subset $B_s$, by tuning the ratio between the number of attack points and the number of data points.
% (removing a number of virtual points)
\begin{algorithm}
\caption{Manifold attack}
\label{algorithm:manifold_attack}
\begin{algorithmic} 
\REQUIRE Data points $\{ \textbf{x}_1,..,\textbf{x}_N\}$, embedding loss $\mathcal{L}_e()$, model $g()$, $\xi$, $n\_iters$, an anchoring rule.
\STATE \textbf{initialize}: $g()$
\STATE Create $M$ sets of anchor points $\{ \mathbf{z}_1^k,..,\mathbf{z}_p^k \},\forall k = 1,..,M$ by the anchoring rule
\FOR{$epoch = 1$ \textbf{to} $n\_epoch$}
% \STATE Take a subset $\mathcal{N}_b$ of $\{1,..,M+N \}$
\STATE Initialize $\gamma^k \in \mathbb{R}^p$ for constraints in \cref{equa:attack_point_with_anchors} 
\STATE $\tilde{\mathbf{x}}^k = \gamma_1^k\mathbf{z}_{1}^k + \gamma_2^k\mathbf{z}_{2}^k + ... + \gamma_p^k\mathbf{z}_{p}^k, \forall k = 1,..,M$ 
\STATE Set $\mathcal{B} = \{g(\tilde{\mathbf{x}}^1),..,g(\tilde{\mathbf{x}}^M)\} \cap \{g(\mathbf{x}_1),..,g(\mathbf{x}_N)\} $ and divide it into subsets $\mathcal{B}_s$
\FOR{each $\mathcal{B}_s$}
\STATE  $L = \sum_{\mathbf{a} \in \mathcal{B}_s } \mathcal{L}_e\big(\mathbf{a}, \mathcal{B}_s \backslash \{ \mathbf{a} \} \big)$
\STATE  Update $\{\tilde{\mathbf{x}}^i | g(\tilde{\mathbf{x}}^i) \in \mathcal{B}_s  \}$ to maximize $L$ by \cref{algorithm:Virtual_point_update}
\STATE  Update $g()$ to minimize $L$ 
\ENDFOR
\ENDFOR
\STATE \textbf{Output: $g()$} 
\end{algorithmic}
\end{algorithm}

\begin{algorithm}
\caption{Virtual points update}
\label{algorithm:Virtual_point_update}
\begin{algorithmic}
\REQUIRE $m$ sets of anchor points $\{ \mathbf{z}_1^k,..,\mathbf{z}_p^k \}$, $\gamma^k$, $\forall k = 1,..,m$, loss $L$, $\xi ,n\_iters$.
\STATE $\tilde{\mathbf{x}}^k = \gamma_1^k\mathbf{z}_{1}^k + \gamma_2^k\mathbf{z}_{2}^k + ... + \gamma_p^k\mathbf{z}_{p}^k, \forall k = 1,..,m$
\FOR{$i = 1$ \textbf{to} $n\_iters$}
\STATE Calculate gradient $\nabla L$ (w.r.t $ [\gamma^1,..,\gamma^{m}]$) of function $L(\tilde{\mathbf{x}}^1,..,\tilde{\mathbf{x}}^m)$  
\STATE  $ [\gamma^1,..,\gamma^{m}] \leftarrow [\gamma^1,..,\gamma^{m}] + \xi \nabla_{\gamma} L(\tilde{\mathbf{x}}^1,..,\tilde{\mathbf{x}}^{m})$
\STATE  $\gamma^k \leftarrow \Pi_{ps}(\gamma^k), \forall k = 1,..,{m}$
\STATE $\tilde{\mathbf{x}}^k = \gamma_1^k\mathbf{z}_{1}^k + \gamma_2^k\mathbf{z}_{2}^k + ... + \gamma_p^k\mathbf{z}_{p}^k, \forall k = 1,..,m$
\ENDFOR
\STATE \textbf{Output: $\tilde{\mathbf{x}}^1,..,\tilde{\mathbf{x}}^{m}$} 
\end{algorithmic}
\end{algorithm}

Algorithm \ref{algorithm:Virtual_point_update} represents the update step for multiple attack points. %The gradient update for $\gamma$ is the simplest version by default and we can replace it with the other one in the gradient-based family. 
%Since the attack stage is executed a lot of times (alternately with model update stage), we recommend to use $n\_iter$ that approximates to less than 5 iterations. 
%Since finding attack points is based on gradient-based method, therefore 
We assumed that the embedding loss $\mathcal{L}_e$ is smooth with respect to $\gamma$ and used a gradient-based algorithm to estimate the latter. Nevertheless, in PGS, there are several methods whose embedding loss is not even continuous. For example, in LLE (\ref{equa:LLE_loss}), the embedding loss takes into account the $k$ nearest neighbors of a point, which might change throughout the estimation of an attack point, producing a discontinuity. %Imagine that after a gradient update step in manifold attack, an attack point changes several neighbors among its actual $k$ nearest neighbors, then the embedding loss for this attack point takes into account new neighbors, which make it discontinuous. 
To circumvent this problem, we use several strategies to avoid singularities:   
\begin{itemize}
    % \item[-] By the anchor rule. We set anchor points in order that a virtual point, which is inside the zone defined by anchor points, does not change its $k$ nearest neighbors.
    \item[-] By reducing the gradient step $\xi$ %or the number of iterations $n\_iter$, 
    which limits the virtual point displacement.
    \item[-] By taking a small number of attack points in each subset $\mathcal{B}_s$ or by using randomly a part of attack points to perform the attack while fixing other attack points, in an attack stage.
    \item[-] By updating $\gamma$ only if embedding loss increases.
\end{itemize}

Besides, some metrics might be approximated by smooth functionals. For instance, in the contrastive loss (\ref{equa:Contrastive_loss}), we can replace the metric $d_x()$ which outputs only 0 or 1, with $d_x(\mathbf{x}_i,\mathbf{x}_j) = \text{exp} \big( \frac{-\norm{\mathbf{x}_i-\mathbf{x}_j}_2^2}{2\sigma_i^2} \big)$ to make embedding loss continuous.
 
\subsection{Pairwise preservation of geometric structure}
\label{subsec:Pairwise_manifold_learning}
For some PGS methods as MDS or LE, the embedding loss $\mathcal{L}_e$ can be decomposed into the sum of elementary pairwise loss $l_e$: 
\begin{equation}
\mathcal{L}_e(\mathbf{a}, \mathcal{B}) = \sum_{\mathbf{b} \in \mathcal{B}} l_e(\mathbf{a}, \mathbf{b}) 
\end{equation}

Then the batch loss $L$ (in \cref{algorithm:manifold_attack}) can be modified into: 
\begin{equation}
L = \sum_{\mathbf{a} \in \mathcal{B}_s } \mathcal{L}_e\big(\mathbf{a}, \mathcal{B}_s \backslash \{ \mathbf{a} \} \big) = \sum_{\mathbf{a} \in \mathcal{B}_s}\sum_{\mathbf{b} \in \mathcal{B}_s,\mathbf{b} \neq \mathbf{a} } l_e(\mathbf{a},\mathbf{b})
\end{equation}

Following this change, $L$ can be decomposed into three parts: 
\begin{alignat*}{2}
&\sum_{\mathbf{a} \in \mathcal{B}_s^d}\sum_{\mathbf{b} \in \mathcal{B}_s^d,\mathbf{b}\neq \mathbf{a}} l_e(\mathbf{a}, \mathbf{b}) \quad \quad \quad &&(\text{data-data}) \\
&\sum_{\mathbf{a} \in \mathcal{B}_s^d}\sum_{\mathbf{b} \in \mathcal{B}_s^v} l_e(\mathbf{a}, \mathbf{b}) &&(\text{data-virtual}) \\
&\sum_{\mathbf{a} \in \mathcal{B}_s^v}\sum_{\mathbf{b} \in \mathcal{B}_s^v,\mathbf{b} \neq \mathbf{a}} l_e(\mathbf{a}, \mathbf{b}) &&(\text{virtual-virtual}),
\end{alignat*}
where $\mathcal{B}_s^d$ and $\mathcal{B}_s^v$ are respectively set that contains all embedded data points (or observed samples) and all embedded virtual points of $\mathcal{B}_s$. 

In some PGS losses which can be decomposed into the sum of elementary pairwise loss, we can balance the effect of the virtual points with respect to the observed samples, not only by tuning the ratio between the number of virtual points and the number of observed samples in $B_s$, but also by weighting each of three parts above which corresponds to the settings observed-observed, observed-virtual and virtual-virtual. %The elementary pairwise loss $l_e$ for each part is not required to be the same, \textit{e.g.} we can use $l_e$ in MDS for data-data and $l_e$ in LE for data-virtual\dots         

\subsection{Settings of anchor points and initialization of virtual points}
\label{subsec:Settings of anchor points}

In this section, we provide two settings (or rules) for computing anchor points with the corresponding initializations. These settings need to be chosen carefully to guarantee that virtual points are on the sample underlying manifold.  

\textbf{Neighbor anchors}: The first anchor point $\mathbf{z}_1$ is taken randomly from $\mathcal{X}$, then the next $(p-1)$ anchor points $\mathbf{z}_2,..,\mathbf{z}_p$ are taken as $(p-1)$ nearest neighbor points of $\mathbf{z}_1$ in $\mathcal{X}$ (Euclidean metric by default). Here, we assume that the convex hull of a sample and its neighbors is likely comprised in the samples manifold. The number of anchors $p$ needs to be small compared to the number of data points $N$. The initialization for virtual points can be set by taking $\gamma_i \sim \mathcal{U}(0,1), \forall i = 1,..,p$ then normalize to have $\sum_{i=1}^p \gamma_i = 1$.   

\textbf{Random anchors}: The second setting is inspired by Mix-up method \cite{ZhangHongyi17mix}. $p$ anchors are taken randomly from $\mathcal{X}$ and we take $\gamma \sim \text{Dirichlet}(\alpha_1,..,\alpha_p)$. If $\alpha_i \ll 1, \forall i = 1,..,p$, the Dirichlet distribution returns $\gamma$ where $\gamma_i \geq 0$, $\sum_{i=1}^p \gamma_i = 1$. In particular, there is a coefficient $\gamma_k$ much greater than other ones with a strong probability, which implies that virtual points are more probably in the neighborhood of a data sample. Since the manifold attack tries to find only local maximum by gradient-based method, if $\xi$ and $n\_iters$ are both small, we expect that attack points in the attack stage do not move too far from their initiated position, remaining on the manifold of data. Note that, in the case $\alpha_i = 1, \forall i = 1,..,p$, the Dirichlet distribution become the Uniform distribution.

To ensure that the coefficient $\gamma_k$ is always much greater than other ones, we apply one more constraint: $\gamma_k \geq \tau$, and by taking $\tau$ close to 1. The constraints in (\ref{equa:attack_point_with_anchors}) become: 
\begin{equation}
\begin{split}
&\gamma_1, \gamma_2,.., \gamma_p \geq 0, \\
& \gamma_1+ \gamma_2+...+\gamma_p = 1, \\
&\gamma_k \geq \tau, (\tau < 1). \\
\end{split}    
\end{equation}

Then the projection in algorithm \ref{algorithm:Projection_for_positive_and_sum} needs to be slightly modified to incorporate this new constraint. We define the projection $\gamma = \Pi_{ps}^{'}(\kappa)$ as follow:
\begin{equation}
\label{equa:projection_sum_constraint_bias}
\begin{split}
& \kappa' \leftarrow \kappa \\
& \kappa'_k \leftarrow \kappa'_k - \tau \\
&\gamma \leftarrow \Pi_{ps}(\kappa', c = 1 - \tau) \\
&\gamma_k \leftarrow \gamma_k + \tau \\
\end{split}
\end{equation}

%In some cases, we can go a little further by applying parameter $s$ that control expansion ($s > 1$ or
%contraction $0 > s > 1$ of the convex envelope as shown in figure \ref{fig:individual manifold attack}.

\section{Applications of manifold attack}
\label{sec:Applications of manifold attack}

We present several applications of manifold attack for NNMs that use PGS task. Firstly, we show advantages of manifold attack for a PGS task when only few training samples are available. Secondly, we show that applying manifold attack \textit{in moderation} improves both generalization and adversarial robustness.

\subsection{Preservation of geometric structure with few training samples}
\label{subsec:Embedded representation}
\begin{figure}[h]
\centering
  \includegraphics[width=0.8\linewidth]{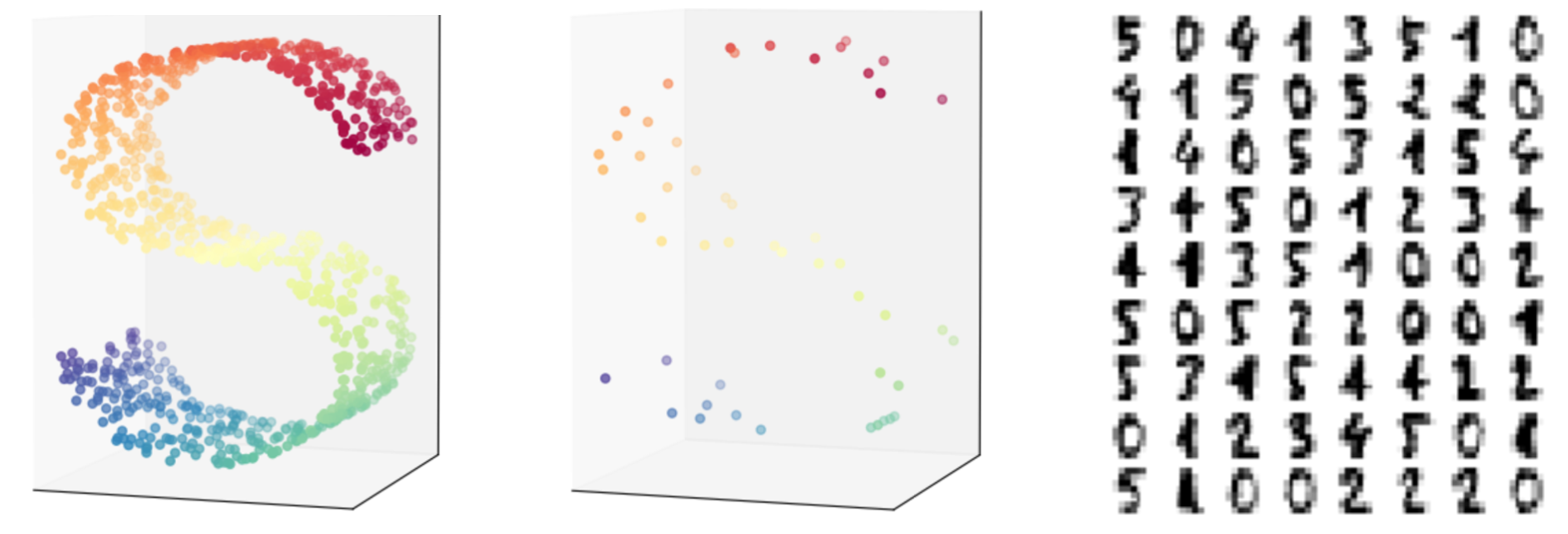}
  \caption{Left: S curve data 1000 samples. Center: S curve data 50 samples. Right: Digit data.}
  \label{fig:S_curve_data}
\end{figure}

For this experiment, we use the S curve data and Digit data.
% are created by the module \textit{datasets} in \textit{scikit-learn}. 
The S curve data contains $N = 1000$ 3-dimensional samples, as shown in figure \ref{fig:S_curve_data}. The Digit data contains $N = 1797$ images, of size $8\times8$ of a digit. We want to compute 2-dimensional embeddings for these data. Each data is separated into two sets: $N_{tr}$ samples are randomly taken for training set and the $N_{te}$ remaining samples are used for testing. %$\mathbf{x}$ denotes a sample, 
$\mathbf{x}^{tr}$ denotes a training sample and $\mathbf{x}^{te}$ denotes a testing sample. We perform four training modes as described in \cref{tab:four_tests_evaluation} with a neural network model $g()$. The evaluation loss, after optimizing model $g()$, is defined as: 
\begin{equation}
L_{ev} = \frac{1}{N_{te}} \sum_{i=1}^{N_{te}}\mathcal{L}_e(g(\mathbf{x}^{te}_i),\{ g(\mathbf{x}^{te}_j) | j \neq i \})
\end{equation}

\begin{table}[h]
\begin{center}
\begin{tabular}{|c|p{12cm}|}
\hline
 & \\[\dimexpr-\normalbaselineskip+0.5em]
\multicolumn{1}{|c|}{ Mode  } & \multicolumn{1}{c|}{Description of objective function} \\
& \\[\dimexpr-\normalbaselineskip+0.5em]
\hline
& \\ [\dimexpr-\normalbaselineskip+0.5em]
1. REF  & PGS that takes into account both training and testing samples: 
\begin{equation*}
 L_{tr} = \frac{1}{N} \sum_{i=1}^{N}\mathcal{L}_e(g(\textbf{x}_i),\{ g(\textbf{x}_j) | j \neq i \})
 \end{equation*}
The result of this training mode is considered as ``reference'' in order to compare to other training modes. \\
& \\[\dimexpr-\normalbaselineskip+0.5em]
\hline
& \\ [\dimexpr-\normalbaselineskip+0.5em]
2. DD & PGS that takes only training (observed) data samples: 
 \begin{equation*}
 L_{tr} =\frac{1}{N_{tr}} \sum_{i=1}^{N_{tr}}\mathcal{L}_e(g(\textbf{x}_i^{tr}),\{ g(\textbf{x}_j^{tr}) | j \neq i \} ).
\end{equation*} \\
& \\[\dimexpr-\normalbaselineskip+0.5em]
\hline
& \\ [\dimexpr-\normalbaselineskip+0.5em]
3. RV & Using virtual points as supplement data, virtual points are only randomly initialized and without attack stage (by setting $n\_iters = 0$ in \cref{algorithm:Virtual_point_update}): 
\begin{equation*}
L_{tr} = \frac{1}{|\mathcal{B}|} \sum_{\textbf{a}_i \in \mathcal{B}}\mathcal{L}_e(\textbf{a}_i,\mathcal{B} \backslash \{ \textbf{a}_i \}),
\end{equation*}
where $\mathcal{B} = \{g(\tilde{\textbf{x}}^1),..,g(\tilde{\textbf{x}}^M)\} \cap \{g(\textbf{x}_1^{tr}),..,g(\textbf{x}_{N_{tr}}^{tr})\} $.  \\  
& \\[\dimexpr-\normalbaselineskip+0.5em]
\hline
& \\ [\dimexpr-\normalbaselineskip+0.5em]
4. MA & Using manifold attack, the same objective function as the previous case, except $n\_iters \neq 0$ (virtual points become attack points). \\
\hline
\end{tabular}
\end{center}
\caption{Four training modes: REF (Reference), DD (Data-Data), RV (Random virtual) and MA (Manifold Attack), and their corresponding objective function.}
\label{tab:four_tests_evaluation} 
\end{table}

The anchoring rule, embedding loss $\mathcal{L}_e$ and the model $g()$ are precised in the following. 

\underline{Anchoring rule.} Two settings are considered: 
\begin{itemize}
    \item[-] \textit{Neighbor anchors} (NA): A set of anchor points is composed by a sample with its 4 nearest neighbors. In this case, we have $M = N_{tr}$ sets of anchor points and $p = 5$ anchor points in each set. The coefficient $\gamma$ is initialized by uniform distribution.
    \item[-] \textit{Random anchors} (RA): We take randomly 2 points among $N_{tr}$ training points to create a set of anchor points. In this case, we have $M = \binom{N_{tr}}{2}$ sets of anchor points and $p=2$ anchor points in each set. The coefficient $\gamma$ is initialized by the Dirichlet distribution with $\alpha_i = 0.5, \forall i = 1,..,p $.
\end{itemize} 

\underline{The embedding loss.} We employ the embedding loss MDS and LE as described in section \ref{sec:Preservation of geometric structure}, with the default metrics. For similarity metric $d_x()$ in LE method, we take $\sigma = 0.2$ for S curve data and $\sigma = 0.5$ for Digit data.

\underline{The model.} A simple structure of CNN is used. Here are the detailed architectures for each CNN as the dimension of the inputs are different for the two datasets:

\begin{itemize}
    \item[-] S curve data: \textit{Conv1d}[$1,4,2$] $\rightarrow$ \textit{ReLu} $\rightarrow$ \textit{Conv1d}[$4,4,2$] $\rightarrow$ \textit{ReLu}$\rightarrow$ \textit{Flatten} $\rightarrow$ \textit{Fc}[$4,2$].
    \item[-] Digit data: \textit{Conv2d}[$1,8,3$] $\rightarrow$ \textit{ReLu} $\rightarrow$ \textit{Conv2d}[$8,16,3$] $\rightarrow$ \textit{ReLu}$\rightarrow$ \textit{Flatten} $\rightarrow$ \textit{Fc}[$64,2$].
\end{itemize} 

For LE method, two additional constraints are imposed to avoid trivial embeddings: 
\begin{equation}
\begin{gathered}
    \mathrm{E}(\mathbf{A}^{tr}) =[\mathrm{E}(\mathbf{A}^{tr}[1,:]),..,\mathrm{E}(\mathbf{A}^{tr}[d,:])]^\top  = \mathbb{0}_{d} \\
    \Sigma(\mathbf{A}^{tr},\mathbf{A}^{tr})  = I_d \\
\end{gathered}
\end{equation}
where $d = 2$ is the number of output dimensions, $\mathbf{A}^{tr} = [\mathbf{a}_1^{tr},..,\mathbf{a}_{N_{tr}}^{tr}] = [g(\mathbf{x}_1^{tr}),..,g(\mathbf{x}_{N_{tr}}^{tr})] $. To adapt these constraints, we add a normalization layer at the end of model $g()$: $(g(\mathbf{x}) -\mathrm{E}(\mathbf{A}^{tr}))\Sigma^{-1}(\mathbf{A}^{tr},\mathbf{A}^{tr})$, where $\Sigma^{-1}$ is performed by Cholesky decomposition.

To simulate the case of few training samples, we fix $N_{tr} = 100 $ for MDS method and $N_{tr} = 50 $ for LE method. The balance between virtual points and samples is controlled by the couple $\lambda =$ (number of virtual points in $\mathcal{B}_s$ , number of observed samples in $\mathcal{B}_s$). We set $\lambda = (2,5)$ for MDS method and $\lambda = (5,10)$ for LE method. The gradient step $\xi$ is selected from $\{0.1,1,10\}$ and the number of iterations is fixed at $n\_iters = 2$. 

The initialization of model $g()$ is impactful, especially since there are few training data. Five different initialization of model for each method are performed. The mean and the standard deviation of the evaluation loss $L_{ev}$ are represented in \cref{tab:MDS and LE S data}. Firstly, we see that using random virtual (RV) points as additional data points gives a better loss than using only data points. %this is evident since we use more samples to train the model. 
Secondly, using manifold attack (MA) further improves the results which shows the benefit of the proposed approach to regularize the model. 

For the S curve data, initialization by \textit{Neighbors anchors} (NA) gives a better result compared to initialization by \textit{Random anchors} (RA). However, for the Digit data, initialization by \textit{Random anchors} gives a better result. This is due to the fact that in the S curve data, \textit{Neighbor Anchors} covers better the manifold of data than \textit{Random Anchor}. On the other hand, in Digit data, \textit{Neighbor Anchors} (by using Euclidean metric to determine nearest neighbors) can generate, with greater probability, a virtual point that is not in the manifold of data. This leads to a greater evaluation loss compared to \textit{Random Anchor}. 

\begin{table}[!ht]
\begin{center}
\rowcolors{2}{gray!20}{white}
\begin{tabular}{|l|c|c|c|c|}
\hline
\multicolumn{1}{|c|}{} &  \multicolumn{2}{c|}{S curve data} & \multicolumn{2}{c|}{Digit data}\\
\hline
\multicolumn{1}{|c|}{Mode / Method} & MDS & LE & MDS & LE \\ [0.5ex]
 \hline
% \rowcolor{lightgray}
REF & $130.7 \pm 24.74  $  & $0.399 \pm 0.07  $ & $2015\pm14$ & $0.07 \pm 0.002 $ \\
 \hline
DD & $352.56 \pm 119.19  $ & $1.21 \pm 0.46  $ &$2409\pm78$ & $0.58 \pm 0.07 $ \\ 
 \hline
%  \rowcolor{lightgray}
RV (NA) & $173.87 \pm 9.38  $ & $0.59 \pm 0.11 $ & $2395\pm73$ &$0.31 \pm 0.03 $ \\
 \hline
MA (NA) & $170.62 \pm 5.89 $ & $0.55 \pm 0.07 $&$2362\pm63$ &$0.24 \pm 0.03 $ \\
\hline
%  \rowcolor{lightgray}
RV (RA) & $183.42 \pm 18.13  $ & $0.65 \pm 0.14 $ &$2342\pm56$  & $0.22 \pm 0.03 $ \\
 \hline
MA (RA) & $169.04 \pm 5.30 $ & $0.63 \pm 0.14 $ & $2331 \pm 56 $&$0.2 \pm 0.02 $ \\
 \hline
\end{tabular}
\end{center}
\caption{\label{tab:MDS and LE S data} Evaluation loss $L_{ev}$ of two PGS methods MDS and LE, in four modes: Reference (REF), Data-Data (DD), Random Virtual (RV), Manifold Attack(MA) (as described in table \ref{tab:four_tests_evaluation}) and two initialization strategies: Neighbor Anchors (NA) and Random Anchors (RA).}
\end{table}

\begin{figure}[!h]
\hspace{-1cm}
  \includegraphics[width=1.1\linewidth]{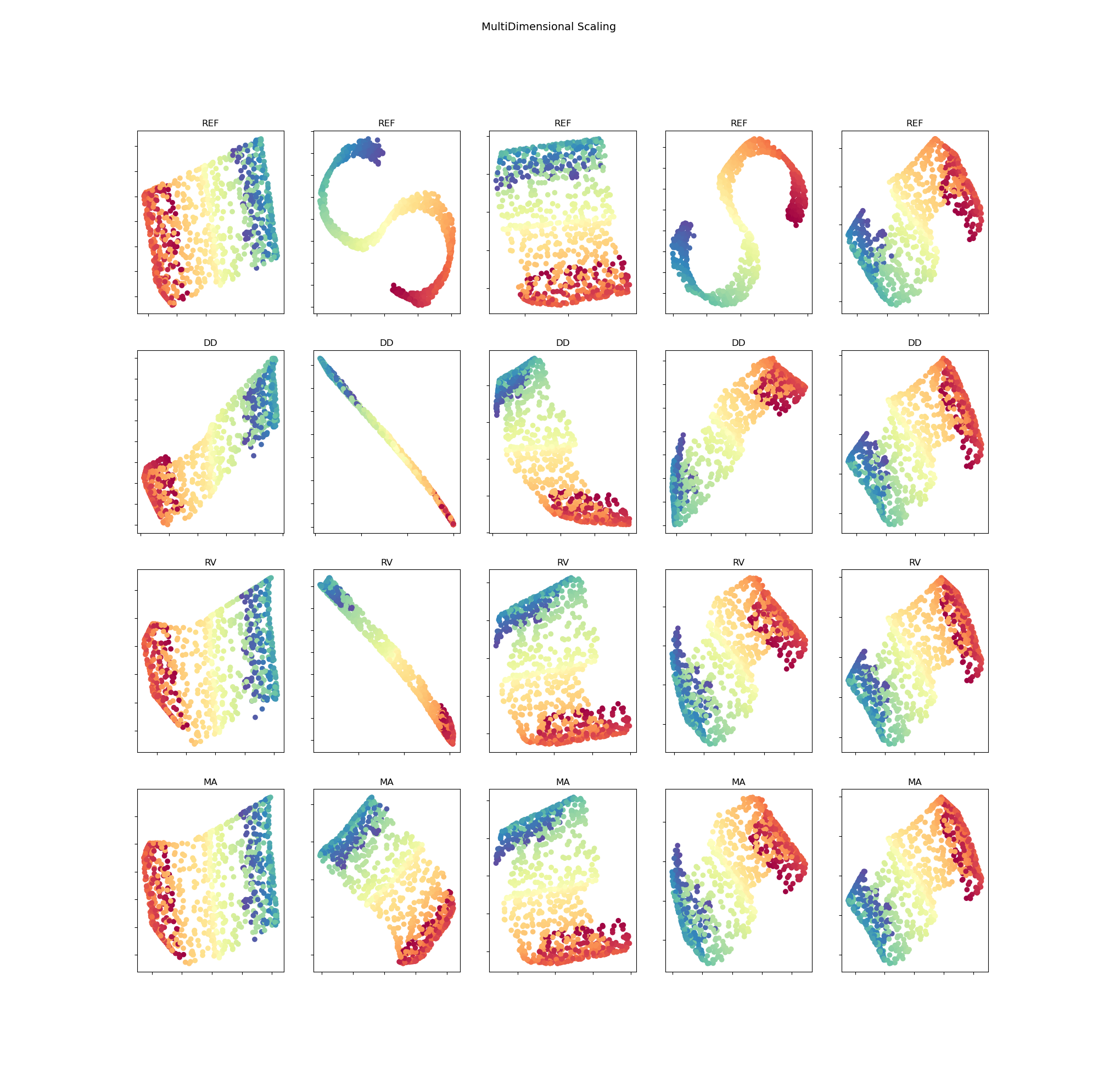}
  \caption{Five evaluations with different initialization of model for the S curve data, using MDS method with four modes: Reference (REF), Data-Data (DD), Random Virtual (RV), Manifold Attack(MA) (as described in table \ref{tab:four_tests_evaluation}) and two initialization strategies: Neighbor Anchors (NA) and Random Anchors (RA). We see clearly the effect of Manifold Attack by the second column. Thus, the embedded representation of Manifold Attack is more spread compared to Random Virtual.}
  \label{fig:MDS_S_data}
\end{figure}

\begin{figure}[!h]
\hspace{-1cm}
  \includegraphics[width=1.1\linewidth]{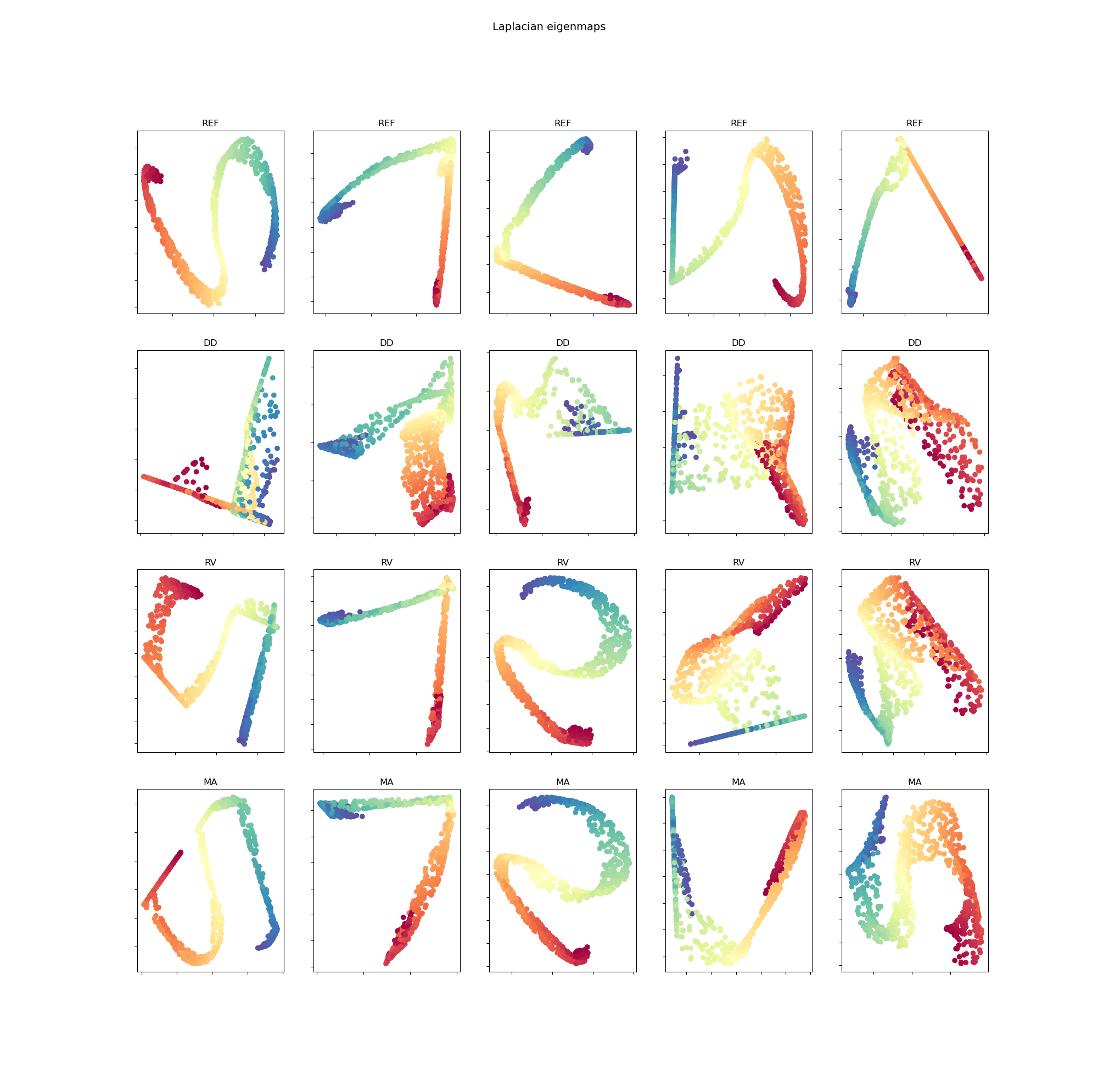}
  \caption{Five evaluations with different initialization of model for the S curve data, using LE method with four modes: Reference (REF), Data-Data (DD), Random Virtual (RV), Manifold Attack(MA) (as described in table \ref{tab:four_tests_evaluation}) and two initialization strategies: Neighbor Anchors (NA) and Random Anchors (RA). We see clearly the effect of Manifold Attack by the fourth column, where the embedded representation shape of Manifold Attack is more similar to Reference than one of Random Virtual.}
  \label{fig:LE_S_data}
\end{figure}

The five embedded representations, respectively with five different initialization of $g()$, for testing samples in S curve data are found in figure \ref{fig:MDS_S_data} for MDS method and in figure \ref{fig:LE_S_data} for LE method.
% We give some comments about the result in the caption. 

\subsection{Robustness to adversarial examples}
\label{subsec:Robustness to adversarial examples}
In this subsection, we combine manifold attack with supervised learning to assess the adversarial robustness of model. Let start with a general objective function which is created by a supervised loss and a PGS loss :
% \begin{equation}
% \begin{aligned}
\begin{align}
\label{equa:supervised manifold attack objective function}
    &\min_{\theta}\max_{\gamma} \left( L_{s} + \lambda L_{pgs}\right) \nonumber \\ 
    = &\min_{\theta}\max_{\gamma} \frac{1}{N_l} \sum_{i=1}^{N_l} \mathcal{L}_c(f_{\theta}(\mathbf{x}^l_i),y_i) + \lambda \mathcal{L}_t\left(f_{\theta}^{(l)}(\mathbf{x}_1^l),..,f_{\theta}^{(l)}(\mathbf{x}_{N_l}^l), f_{\theta}^{(l)}(\tilde{\mathbf{x}}^1),..,f_{\theta}^{(l)}(\tilde{\mathbf{x}}^M) \right)
% \end{aligned}
% \end{equation}
\end{align}

where $(\mathbf{x}^l_i,y_i)$ means a sample and its corresponding label. $\mathcal{L}_c$ is a dissimilarity metric, like for instance the Cross Entropy. $\mathcal{L}_t$ is a PGS loss that includes eventually all observed samples $\mathbf{x}$ and virtual points $\tilde{\mathbf{x}}$. Note that the coordinates $\gamma$ of all virtual points are constrained by (\ref{equa:attack_point_with_anchors}).

In the following, we construct $\mathcal{L}_t$ by a particular PGS loss, called Mix-up manifold learning loss \cite{Verma19ICT, Verma2019ManifoldMixUp, ZhangHongyi17mix} :

\begin{equation}
\label{equa:mix-up general}
\begin{split}
&l_{mu} \left(f_{\theta}^{(l)}(\tilde{\mathbf{x}}),f_{\theta}^{(l)}(\mathbf{x}_i),f_{\theta}^{(l)}(\mathbf{x}_j)\right) \\
= &\mathcal{L}_c\left(f_{\theta}^{(l)}(\gamma_1\mathbf{x}_i + \gamma_2\mathbf{x}_j),\gamma_{1}f_{\theta}^{(l)}(\mathbf{x}_i) + \gamma_{2}f_{\theta}^{(l)}(\mathbf{x}_j)\right)\\
\text{where: } & \gamma_1 + \gamma_2 = 1 \text{ and } \gamma_1 \sim \text{Beta}(\alpha,\alpha), 0 < \alpha \ll 1 
\end{split}
\end{equation}

In this configuration, the virtual point $\tilde{\mathbf{x}} = \gamma_1\mathbf{x}_i + \gamma_2\mathbf{x}_j$, where $\mathbf{x}_i$ and $\mathbf{x}_j$ are two anchor points. The distribution Beta is just a particular case of the Dirichlet distribution where the number of anchor points $p=2$ (in \cref{subsec:Settings of anchor points}). In case of supervised learning, we perform PGS between the original samples representation and the final NNM output (which can be considered as a latent representation). Thus, we implement a slightly modified version of equation (\ref{equa:mix-up general}):  
 
\begin{equation}
\label{equa:mix-up labelled}
\begin{split}
&l_{mu} \left(f_{\theta}(\tilde{\mathbf{x}}),f_{\theta}(\mathbf{x}^l_i),f_{\theta}(\mathbf{x}^l_j)\right) \\
= & \mathcal{L}_c\left(f_{\theta}(\gamma_1\mathbf{x}_i^l + \gamma_2\mathbf{x}_j^l),\gamma_{1}y_i + \gamma_{2}y_j\right)\\
\text{where: } & \gamma_1 + \gamma_2 = 1 \text{ and } \gamma_1 \sim \text{Beta}(\alpha,\alpha), 0 < \alpha \ll 1  
\end{split}    
\end{equation}

Then, we replace this explicit PGS loss in equation (\ref{equa:supervised manifold attack objective function}) and take $\lambda = 1$:

\begin{align}
\label{equa:Mix-up adversarial}
     \min_{\theta}\max_{\gamma} \frac{1}{N_l} \sum_{i=1}^{N_l}&\mathcal{L}_c(f_{\theta}(\mathbf{x}^l_i),y_i) + \lambda \frac{1}{N_l^2N_{\gamma}}\sum_{k=1}^{N_{\lambda}}\sum_{i=1}^{N_l}\sum_{\substack{j=1 \\ j\neq i}}^{N_l} \mathcal{L}_c\left(f_{\theta}(\gamma_1\mathbf{x}_i^l + \gamma_2\mathbf{x}_j^l),\gamma_{1}y_i + \gamma_{2}y_j\right) \\
     =\min_{\theta}\max_{\gamma} &\frac{1}{N_l^2N_{\gamma}}\sum_{k=1}^{N_{\lambda}}\sum_{i=1}^{N_l}\sum_{j=1}^{N_l} \mathcal{L}_c\left(f_{\theta}(\gamma_1\mathbf{x}_i^l + \gamma_2\mathbf{x}_j^l),\gamma_{1}y_i + \gamma_{2}y_j\right)\\
     \text{where: } & \gamma_1 + \gamma_2 = 1 \text{ and } \gamma_1 \sim \text{Beta}(\alpha,\alpha), 0 < \alpha \ll 1  \\
     & N_\gamma \text{ is the number of samplings $\gamma$ from } \text{Beta}(\alpha,\alpha).
\end{align}

We call problem (\ref{equa:Mix-up adversarial}) adversarial Mix-up because this problem is developed from Mix-up\cite{ZhangHongyi17mix}, by adding an adversarial learning for $\gamma$. In practice, to optimize the problem (\ref{equa:Mix-up adversarial}), we perform an attack stage (algorithm \ref{algorithm:Virtual_point_update}) to find $\gamma$ that gives the greater PGS loss, before performing model update stage for $f_{\theta}()$. We repeat alternatively these two stages until convergence. Note that, we can take $\gamma_2 = 1-\gamma_1$, so that we only need to deal with one variable $\gamma_1$ to maximize PGS loss. The projection (\ref{equa:prox_operation_for_positif_and_constraint}) for $\gamma_1$ is now just the clamping function, to make sure that $\gamma_1$ is between 0 and 1.

We compare four supervised training methods: ERM (Empirical Risk Minimization, which is thus classical supervised learning with Cross Entropy loss), Mix-up, Adversarial Mix-up and Cut-Mix \cite{YunSangdoo19} on ImageNet dataset with the model ResNet-50 \cite{HeKaiming15}, which has about 25.8M trainable parameters. We use ImageNet dataset. We retrieve 948 classes consisting in 400 labelled training samples and 50 testing samples to evaluate the models.

% \begin{table}[!ht]
% \begin{center}
% \rowcolors{2}{gray!20}{white}
% \begin{tabular}{|p{65mm}|P{10mm}|P{10mm}|P{15mm}|P{15mm}|}
% \hline
% \multicolumn{1}{|c|}{Method / Data evaluation} & \multicolumn{2}{c|}{Testing set} & \multicolumn{2}{c|}{Adversarial examples}  \\
% \hline
% & Top-1 &Top-5 &  Top-1  & Top-5 \\
% \hline
% ERM  & 33.84 & 12.46 & 81.69  & 59.14\\
% \hline
% Mix-up \cite{ZhangHongyi17mix} & 32.13 & 11.35 & 75.57 & 49.41 \\
% \hline
% Mix-up Adversarial (1) ($\xi=0.01 \rightarrow 0.001$) & 31.82 & 11.15 & 70.82 & 43.96\\
% \hline
% Mix-up Adversarial (2) ($\xi=0.01$) & 31.35 & 11.03 & 68.30 & 41.18\\
% \hline
% Mix-up Adversarial (3) ($\xi=0.1 \rightarrow 0.01$) & 32.57 & 10.98 & 63.62 & 36.15\\
% \hline
% Mix-up Adversarial (4) ($\xi=0.1$) & 31.57 & 10.95 & 66.53 & 39.10\\
% \hline
% Cut-Mix \cite{YunSangdoo19} &30.94 & 10.41 & 81.24 & 58.72\\
% \hline

% \end{tabular}
% \end{center}
% \caption{\label{tab:mix_up_attack}: ImageNet error rate (top-1 and top-5 in \%) on testing set and on adversarial examples on different training modes: ERM, Mix-Up, Mix-up Adversarial and Cut-Mix.}
% \end{table}

\begin{table}[!ht]
\begin{center}
\rowcolors{2}{gray!20}{white}
\begin{tabular}{|p{65mm}|P{10mm}|P{10mm}|P{15mm}|P{15mm}|}
\hline
\multicolumn{1}{|c|}{Method / Data evaluation} & \multicolumn{2}{c|}{Testing set} & \multicolumn{2}{c|}{Adversarial examples}  \\
\hline
& Top-1 &Top-5 &  Top-1  & Top-5 \\
\hline
ERM  & 33.84 & 12.46 & 81.69  & 59.14\\
\hline
Mix-up \cite{ZhangHongyi17mix} & 32.13 & 11.35 & 75.57 & 49.41 \\
\hline
Mix-up Adversarial (1) ($\xi=0.1 \rightarrow 0.01$) & 32.57 & 10.98 & 63.62 & 36.15\\
\hline
Adversarial Mix-up (2) ($\xi=0.01 \rightarrow 0.001$) & 31.82 & 11.15 & 70.82 & 43.96\\
\hline
Cut-Mix \cite{YunSangdoo19} &30.94 & 10.41 & 81.24 & 58.72\\
\hline

\end{tabular}
\end{center}
\caption{\label{tab:mix_up_attack}: ImageNet error rate (top-1 and top-5 in \%) on testing set and on adversarial examples on different training modes: ERM, Mix-Up, Mix-up Adversarial and Cut-Mix.}
\end{table}

We evaluate the error rate for testing set at the end of each epoch and report the best best error rate (top-1 and top-5) in table \ref{tab:mix_up_attack}. We create adversarial examples using Fast Gradient Sign Method (FGSM) \cite{Goodfellow14explaining}, on another trained ERM model, with $\epsilon = 0.05$. In Adversarial Mix-up, $n\_iters$ is fixed at 1. The attack stage is parametrized by the gradient step $\xi$, which is set up following two configurations, (1) $\xi$ is reduced linearly from 0.1 to 0.01 and (2) $\xi$ is reduced linearly from 0.01 to 0.001. Following original articles, $\alpha$ is set at 0.2 for Mix-Up and Adversarial Mix-up and $\alpha=1$ for Cut-Mix. More details for hyper-parameters can be found in \nameref{sec:Appendix}.

Firstly, Mix-up \cite{ZhangHongyi17mix} is a combination between supervised learning task and PGS task loss and it gives a better error rates for both testing set and adversarial examples compared to ERM, which is a standard supervised learning task. %This observation shows the effect of PGS by Mix-up manifold learning. 

Secondly, in Adversarial Mix-up (1) and (2), a trade-off has to be found between error rate for testing set and error rate for adversarial examples. For \textit{strong values} of $\xi$ (1), the error rate for testing sample can be about 0.5\% worse than Mix-up (without using attack stage), but it gains about 13\% for the robustness against adversarial examples. On the other hand, if $\xi$ takes \textit{moderate values} as in (2), error rates for both testing sample and adversarial examples are smaller than those of Mix-Up, but it gains only about 5\% for the robustness against adversarial examples. These two observations show the effect of manifold attack, which is an improved PGS procedure. We conclude that manifold attack not only improves generalization but also significantly improves adversarial robustness of the model.

Finally, Adversarial Mix-up Adversarial (2) provides a slightly worse error rate than Cut-Mix (about 1 \% in the case of testing sample), but it gains about 10\% in the case of adversarial examples.     

%Here, we consider a configuration of only two anchor points $(p=2)$ instead of the configuration of more than two anchor points $(p>2)$. This is because in the latter configuration, we must also favor the fact that the virtual is close to one of anchor point by the distribution Dirichlet, therefore, the configuration of more than two anchor points is probably equivalent to a multi configurations of two anchor points.

It is worth noting that in attack stage, the model $g()$ needs to be differentiable. Thus, in the case of using NNMs with \textit{Dropout} layer for example , the active connections need to be fixed during an attack stage.

\subsection{Semi-supervised manifold attack}
\label{subsec:Semi-supervised manifold attack}

In this subsection, manifold attack is applied to reinforce semi-supervised neural network models that use PGS as regularization. The problem (\ref{equa:supervised manifold attack objective function}) is extended to include unlabelled samples. We assume that the training set contains $N$ samples $\mathbf{x}$, the $N_l$ first samples $\mathbf{x}^l_i$ are labelled and the remaining $N_u = N-N_l$ samples $\mathbf{x}^u_i$ are unlabelled. Hence, the objective function for a semi-supervised manifold attack method: 

\begin{equation}
\label{equa:semi-supervised manifold attack objective function}
\begin{split}
    &\min_{\theta}\max_{\gamma} \left( L_{s}  + \lambda L_{pgs} + \beta L_{u} \right )\\ 
    = &\min_{\theta}\max_{\gamma} \frac{1}{N_l} \sum_{i=1}^{N_l} \mathcal{L}_c(f_{\theta}(\mathbf{x}^l_i),y_i) + \lambda \mathcal{L}_t\left(f_{\theta}^{(l)}(\mathbf{x}_1),..,f_{\theta}^{(l)}(\mathbf{x}_{N}), f_{\theta}^{(l)}(\tilde{\mathbf{x}}^1),..,f_{\theta}^{(l)}(\tilde{\mathbf{x}}^M) +\beta L_{u} \right)
\end{split}    
\end{equation}
where $L_{u}$ refers to other possible unsupervised losses such as pseudo-label, self-supervised learning, etc. 

In the following, we explicit the terms of the problem (\ref{equa:semi-supervised manifold attack objective function}). We build upon MixMatch \cite{Berthelot19MixMatch}, a semi-supervised learning method in the current state-of-the-art. MixMatch use Mix-up manifold learning loss for PGS. Secondly, we add an adversarial learning for $\gamma$ in MixMatch. % to form a concrete case of problem (\ref{equa:semi-supervised manifold attack objective function}), called 
Hence we refer to the resulting instance of \ref{equa:semi-supervised manifold attack objective function} as Adversarial MixMatch. Note that, $\gamma_1$ in MixMatch is slightly different from Mix-up as: 

\begin{equation}
\gamma_1 + \gamma_2 = 1, \gamma_1 \sim \text{Beta}(\alpha,\alpha) \text{ and } \gamma_1 \geq \gamma_2
\end{equation}

Then the projection for $\gamma_1$ is now the clamping function between 0.5 and 1. When two separated variables $\gamma_1$ and $\gamma_2$ are defined, we can use the projector $\Pi_{ps}^{'}$ (\ref{equa:projection_sum_constraint_bias}) defined in subsection \ref{subsec:Settings of anchor points}. We use the \textit{Pytorch} implementation for MixMatch by Yui \cite{Yui19MixMatchImplementation} (with all hyper-parameters by default), then we introduce attack stages, with the number of iterations $n\_iters = 1$. In each experiment, for both CIFAR-10 and SVHN dataset, we divide the training set into three parts: labelled set, unlabelled set and validation set. The number of validation samples is fixed at 5000. The number of labelled samples is 250, and the remaining samples are part of the unlabelled set. We repeat the experiment four times, with different samplings of labelled samples, unlabelled samples, validation samples and different initialization of model Wide ResNet-28 \cite{Zagoruyko2016wideResNet} which has about 1.47M trainable parameters. More details for hyper-parameters can be found in \nameref{sec:Appendix}.  

\begin{table}[!ht]
\begin{center}
\rowcolors{2}{gray!20}{white}
\begin{tabular}{|c|l|c|c|c|c|c|}
\hline
Data & Method / Test  & 1 & 2 & 3 & 4 & Mean\\ 
\hline
CIFAR-10  & MixMatch  & 10.62  & 12.72  & 12.02 & 15.26 & $12.65 \pm 1.68$ \\ 
\hline
CIFAR-10 & MixMatch Adersarial & 8.84 & 10.46 & 10.09 & 12.89 & $10.57 \pm 1.47 $ \\ 
\hline
SVHN & MixMatch  & 6.0925  & 6.73 & 7.802 & 7.37 & $7.0 \pm 0.65 $ \\ 
\hline
SVHN & MixMatch Adersarial & 5.07 & 5.93 &  5.42 & 5.47 & $5.47 \pm 0.3$\\ 
\hline
\end{tabular}
\end{center}
\caption{\label{tab:MixMatchAttack}: CIFAR-10 and SVHN Error rate in four different configurations, each configuration consists of data partitioning and initialization of model parameters. The number of labelled samples is fixed at 250 and the used model is Wide ResNet-28.}
\end{table}

The error rate on testing set, which corresponds to the best validation error rate, is reported in \cref{tab:MixMatchAttack}, for both MixMatch and MixMatch Adersarial. We see that MixMatch Adersarial improves the performance of MixMatch, about $1.5\%$ less on error rate. There is a considerable difference between the error rate of MixMatch reproduced by our experiments and the one reported from the official paper, which might come from the sampling, the initialization of model, the library used (Pytorch \textit{vs} TensorFlow) and the computation of the error rate (error rate associated to best validation error \textit{vs} the median error rate of the last 20 checkpoints). %Note that, there are also other semi-supervised methods in Mix-up family as: Real Mix \cite{nair2019realmix}, EnAET \cite{Wang19EnAET}, ReMixMatch \cite{berthelot2019remixmatch} and we can also apply manifold attack on them to get their attack versions, which are expected to improve the performance.  

\begin{table}[!ht]
\begin{center}
\begin{tabular}{|l|c|c|}
\hline
Method / Data & CIFAR-10 & SVHN \\ 
 \hline
Pi Model $^\diamond$ \cite{Laine16}  & $53.02 \pm 2.05 $ & $17.56 \pm 0.275 $  \\
 \hline
Pseudo Label $^\diamond$ \cite{Lee13PseudoLabel} & $49.98 \pm 1.17 $ & $21.16 \pm 0.88 $ \\
 \hline
VAT $^\diamond$ \cite{Miyato17VAT} & $36.03 \pm 2.82 $ & $8.41 \pm 1.01 $ \\
\hline
SESEMI SSL \cite{Tran19SESEMI} &  & $8.32 \pm 0.13$ \\
\hline
Mean Teacher $^\diamond$ \cite{Tarvainen17MeanTeacher} & $47.32 \pm 4.71 $ & $6.45 \pm 2.43 $ \\
\hline
Dual Student \cite{Ke19DualStudent} & & $4.24 \pm 0.10 $ \\
\hline
MixMatch $^\diamond$ \cite{Berthelot19MixMatch} & $11.08 \pm 0.87 $ &  $3.78 \pm 0.26 $ \\
\hline
MixMatch * \cite{Berthelot19MixMatch} & $12.65 \pm 1.68 $ & $7.0 \pm 0.65 $ \\
\hline
MixMatch Adersarial * & $10.57 \pm 1.47 $ & $5.47 \pm 0.3 $ \\
\hline
Real Mix \cite{nair2019realmix} & $9.79 \pm 0.75 $ & $3.53 \pm 0.38 $ \\
\hline
EnAET \cite{Wang19EnAET} & $7.6 \pm 0.34 $ & $3.21 \pm 0.21 $ \\
\hline
ReMixMatch \cite{berthelot2019remixmatch} & $6.27 \pm 0.34 $ & $3.10 \pm 0.50 $ \\
\hline
Fix Match \cite{Sohn20FixMatch} & $5.07 \pm 0.33 $ & $2.48 \pm 0.38 $ \\
\hline

\end{tabular}
\end{center}
\caption{CIFAR-10 and SVHN error rate of different semi-supervised learning methods. The number of labelled sample is fixed at 250. ($^\diamond$) means that the results are reported from
\cite{Berthelot19MixMatch}. (*) means that that the results are reported from our experiments. The resting results are reported from their corresponding official paper.}
\label{tab:semi_sup_attack_250}
\end{table}

Table \ref{tab:semi_sup_attack_250} shows error rates among semi-supervised methods based on NNMs, for both CIFAR-10 and SVHN dataset with only 250 labelled samples. %Note that, this comparison is really relative because the configuration for dataset, value for hyper-parameters, architecture of used model... are not the same in each experiment. 
We refer also readers to the site PapersWithCode that provides the lasted record for each dataset: CIFAR-10\footnote{CIFAR-10 \url{https://paperswithcode.com/sota/semi-supervised-image-classification-on-cifar-6}} and SVHN \footnote{SVHN \url{https://paperswithcode.com/sota/semi-supervised-image-classification-on-svhn-1}}.

\section{Conclusion}
\label{sec:Conclusions}

\begin{table}[h]
\centering
\begin{tabular}{>{\centering\arraybackslash}m{3.8cm}
    |>{\centering\arraybackslash}m{5cm}
    |>{\centering\arraybackslash}m{4.5cm}}

 & Adversarial Example & Manifold Attack \\
&&\\[0em]
\hline
\begin{tabular}{@{}c@{}}Illustration, \\ Red point: a sample \\ Blue line: border of \\ feasible zone  \end{tabular} & {\begin{tikzpicture}
\draw (2.5,0.5);
\draw (2.5,-2.5);
\draw (-2.5,-2.5);
\draw (-2.5,0.5);
\draw[blue] (0,0) circle (8pt);
\draw[blue] (-2,-2) circle (8pt);
\draw[blue] (2,-2) circle (8pt);
\draw[red,fill=red] (0,0) circle (1pt);
\draw[red,fill=red] (-2,-2) circle (1pt);
\draw[red,fill=red] (2,-2) circle (1pt);
\end{tikzpicture}} & {\begin{tikzpicture}
\draw (2.25,0.5);
\draw (2.25,-2.25);
\draw (-2.25,-2.25);
\draw (-2.25,0.5);
\draw[blue] (0,0) -- (-2,-2) -- (2,-2) -- (0,0);
\draw[red,fill=red] (0,0) circle (1pt);
\draw[red,fill=red] (-2,-2) circle (1pt);
\draw[red,fill=red] (2,-2) circle (1pt);
\end{tikzpicture}} \\
\hline
&&\\[0em]
Feasible zone & Locality of each sample & Convex hull \\
&&\\[0em]
\hline
&&\\[0em]
Variable & Noise $\epsilon$ that has the same size as samples & $\gamma$ has the size which equals to the number of anchor points\\
&&\\[0em]
\hline
&&\\[0em]
PGS task & Points in the locality of a sample must have a similar embedded representation
 & Available for almost PGS tasks
 \\
&&\\
\hline
\end{tabular}
\medskip
\caption{A simple comparison between adversarial examples and manifold attack in general. Note that, in manifold attack, the feasible zone is the whole convex hull in the case of Nearest Anchors (subsection \ref{subsec:Settings of anchor points}). Otherwise, in the case of Random Anchors with the distribution Dirichlet, the feasible zone is rather the neighborhood of each anchor in the convex hull than the center.}
\label{tab:VAT_MA_Comparison} 
\end{table}

We introduced manifold attack as an improved PGS procedure. Firstly, it is more general than adversarial examples (see for a comparison in table \ref{tab:VAT_MA_Comparison}). Secondly, we confirm empirically a statement from \cite{ZhangHongyi17mix}: by using Mix-up as PGS combined with supervised loss, we enhance generalization and improve significantly adversarial robustness. Thirdly, we show that applying manifold attack on Mix-up enhances further generalization and adversarial robustness. There is a trade-off to be found between generalization and adversarial robustness. However, in our experiments, in the 'worst case', we only lose about 1\% in generalization for a gain of about 13\% in adversarial robustness. 

\begin{figure}[h]
\begin{center}
\includegraphics[width=0.7\linewidth]{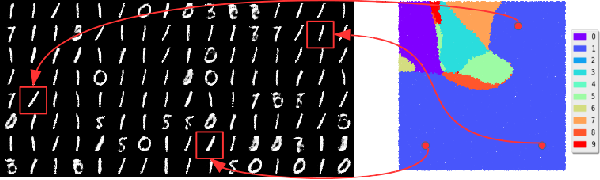}
\end{center}
\caption{Mode collapse observed in a GAN with the MNIST dataset. Left: Samples generated by the generator in GAN. Right: The latent representation of GAN that follows uniform distribution. In this example, the class 1 dominates in the samples generated. Credit: \cite{Tran2018DistGan}.}
\label{fig:GAN_mode_collapse}
\end{figure}

\begin{figure}[h]
\begin{center}
\begin{tikzpicture}

	\node[circle, draw, thick] (z) {$z$};
	\node[circle, draw, thick, right=5em of z] (x) {$x_{fake}$};
	\draw[-stealth, thick] (z) -- node[above] {$G(z)$} node[below] {generator} (x);
	\node[left=of z] (i) {};
	\draw[-stealth, thick] (i) -- node[above] {$p(z)$} (z);
	\node[above=of x, circle, draw, thick] (xt) {$x_{real}$};
	\node[left=5em of xt] (it) {};
	\draw[-stealth, thick] (it) -- node[above] {$p(x)$} (xt);
	\node[circle, draw, thick, right=5em of x, yshift=2.5em] (D) {$x$};
	\node[right=7em of D] (out) {real?};
	\draw[-stealth, thick] (D) -- node[above] {$D(x)$} node[below] {discriminator} (out);
			
	\node[right=2.5em of x, circle, fill, inner sep=0.15em] (pt1) {};
	\node[right=2.5em of xt, circle, fill, inner sep=0.15em] (pt2) {};
			
	\draw[dashed, thick] (pt1) edge[bend left] (pt2);
			
	\node[circle, draw, thick, fill=white, inner sep=0.15em] at ([xshift=-0.9em, yshift=4em]pt1.north) (pt3) {};
			
	\draw[-stealth, thick] (x) -- (pt1);
	\draw[-stealth, thick] (xt) -- (pt2);
	\draw[-stealth, thick] (pt3) -- (D);

    \draw [in=-90,out=-90] (z.south) to (x.south);
    \node[below right = 1 and 0. of z] (ml) {manifold attack};

\end{tikzpicture}
\end{center}
\caption{An GAN that we apply PGS from the latent representation back to original representation.}
\label{fig:GAN_with_PGS}
\end{figure}
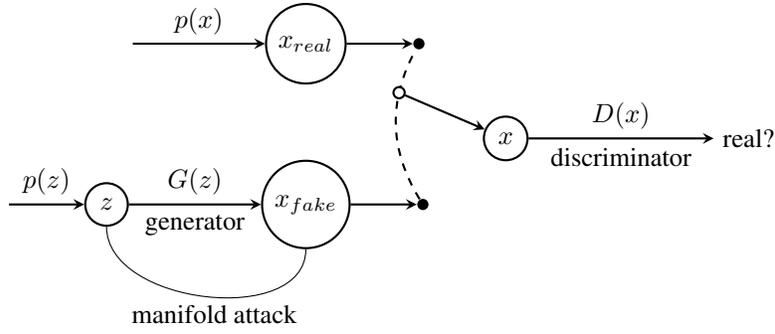

To further improve our method, several directions could be investigated:
\begin{itemize}
    \item Optimization of the layers between those the PGS needs to be applied. Indeed, we could also implement PGS between one latent representation and another one in NNMs as in \cite{Verma2019ManifoldMixUp}.
     
    \item \textit{Mode collapse} is a popular problem while training GAN models. GAN with samples that are well balanced among classes, generated samples by the generator are biased on only a few classes (as showed in figure \ref{fig:GAN_mode_collapse}). This is because the latent representation is not well regularized. We expect to overcome the problem Mode collapse by introducing manifold attack from the latent representation back to the original representation as show in figure \ref{fig:GAN_with_PGS} and by optimizing problem \ref{equa:GAN_manifold_attack}, where $\mathcal{L}_t$ is a PGS task as showed in section \ref{sec:Preservation of geometric structure}.
    
    %\item We expect to enhance the adversarial robustness of model $f_{\theta}$ by applying on-manifold examples to perform PGS. Firstly, the latent representation of data $z$ is learned by a GAN model, with the generator $G$. Then, PGS is applied from the latent representation $z$ (of GAN) to a latent representation of model $f_{\theta}$ as showed in problem \ref{equa:on_manifold_robustness}.
    
\end{itemize}

\begin{equation}
\label{equa:GAN_manifold_attack}
    \min_{G}\max_{D}\max_{z_1,..,z_M \in [0,1]^p}\Big(\amsmathbb{E}_{x\sim p(x)}[\log D(x)] + \amsmathbb{E}_{z \sim p(z)}[1-\log D(G(z))] + \lambda\mathcal{L}_{t}\big(G(z_1),..,G(z_M)\big) \Big)
\end{equation}

% \begin{equation}
% \label{equa:on_manifold_robustness}
%     \min_{\theta}\max_{z_1,..,z_M \in [0,1]^p}\Big(\frac{1}{N_l} \sum_{i=1}^{N_l}\mathcal{L}_c(f_{\theta}(\mathbf{x}^l_i),y_i) + \lambda\mathcal{L}_{t}\big(f_{\theta}^l(G(z_1)),..,f_{\theta}^l(G(z_M))\big) \Big)
% \end{equation}

\section*{Acknowledgement}
This research is supported by the European Community through the grant DEDALE (contract no. 665044) and the Cross-Disciplinary Program on Numerical Simulation of CEA, the French Alternative Energies and Atomic Energy Commission.

\clearpage
\bibliographystyle{unsrt}
\bibliography{references}

\begin{thebibliography}{10}

\bibitem{LeCun15DL}
Yann Lecun, Yoshua Bengio, and Geoffrey Hinton.
\newblock Deep learning.
\newblock {\em Nature Cell Biology}, 521(7553):436--444, may 2015.

\bibitem{Fukushima80}
K~Fukushima.
\newblock Neocognitron: a self organizing neural network model for a mechanism
  of pattern recognition unaffected by shift in position.
\newblock {\em Biological cybernetics}, 36(4):193—202, 1980.

\bibitem{Lecun98}
Y.~{Lecun}, L.~{Bottou}, Y.~{Bengio}, and P.~{Haffner}.
\newblock Gradient-based learning applied to document recognition.
\newblock {\em Proceedings of the IEEE}, 86(11):2278--2324, 1998.

\bibitem{Krizhevsky12}
Alex Krizhevsky, Ilya Sutskever, and Geoffrey~E Hinton.
\newblock Imagenet classification with deep convolutional neural networks.
\newblock In F.~Pereira, C.~J.~C. Burges, L.~Bottou, and K.~Q. Weinberger,
  editors, {\em Advances in Neural Information Processing Systems 25}, pages
  1097--1105. Curran Associates, Inc., 2012.

\bibitem{simonyan2014deep}
Karen Simonyan and Andrew Zisserman.
\newblock Very deep convolutional networks for large-scale image recognition,
  2014.

\bibitem{HeKaiming15}
Kaiming He, Xiangyu Zhang, Shaoqing Ren, and Jian Sun.
\newblock Deep residual learning for image recognition.
\newblock {\em CoRR}, abs/1512.03385, 2015.

\bibitem{Szegedy14}
Christian Szegedy, Wei Liu, Yangqing Jia, Pierre Sermanet, Scott~E. Reed,
  Dragomir Anguelov, Dumitru Erhan, Vincent Vanhoucke, and Andrew Rabinovich.
\newblock Going deeper with convolutions.
\newblock {\em CoRR}, abs/1409.4842, 2014.

\bibitem{Hinton12SpeechRe}
G.~{Hinton}, L.~{Deng}, D.~{Yu}, G.~E. {Dahl}, A.~{Mohamed}, N.~{Jaitly},
  A.~{Senior}, V.~{Vanhoucke}, P.~{Nguyen}, T.~N. {Sainath}, and
  B.~{Kingsbury}.
\newblock Deep neural networks for acoustic modeling in speech recognition: The
  shared views of four research groups.
\newblock {\em IEEE Signal Processing Magazine}, 29(6):82--97, 2012.

\bibitem{Weston08}
Jason Weston and Frédéric Ratle.
\newblock Deep learning via semi-supervised embedding.
\newblock In {\em International Conference on Machine Learning}, 2008.

\bibitem{Goodfellow14explaining}
Ian~J. Goodfellow, Jonathon Shlens, and Christian Szegedy.
\newblock Explaining and harnessing adversarial examples, 2014.

\bibitem{Miyato17VAT}
Takeru Miyato, Shin ichi Maeda, Masanori Koyama, and Shin Ishii.
\newblock Virtual adversarial training: A regularization method for supervised
  and semi-supervised learning, 2017.

\bibitem{Su2019IsRobustnessCostAccuracy}
Dong Su, Huan Zhang, Hongge Chen, Jinfeng Yi, Pin-Yu Chen, and Yupeng Gao.
\newblock Is robustness the cost of accuracy? -- a comprehensive study on the
  robustness of 18 deep image classification models, 2019.

\bibitem{Tsipras2019RobustnessAtOddAccuracy}
Dimitris Tsipras, Shibani Santurkar, Logan Engstrom, Alexander Turner, and
  Aleksander Madry.
\newblock Robustness may be at odds with accuracy, 2019.

\bibitem{Song2018constructing}
Yang Song, Rui Shu, Nate Kushman, and Stefano Ermon.
\newblock Constructing unrestricted adversarial examples with generative
  models, 2018.

\bibitem{Stutz2019RobustnessAndGeneralization}
David Stutz, Matthias Hein, and Bernt Schiele.
\newblock Disentangling adversarial robustness and generalization, 2019.

\bibitem{Dia2019Semantics}
Ousmane~Amadou Dia, Elnaz Barshan, and Reza Babanezhad.
\newblock Semantics preserving adversarial learning, 2019.

\bibitem{Kingma13VAE}
Diederik~P Kingma and Max Welling.
\newblock Auto-encoding variational bayes, 2013.

\bibitem{Goodfellow14GAN}
Ian~J. Goodfellow, Jean Pouget-Abadie, Mehdi Mirza, Bing Xu, David
  Warde-Farley, Sherjil Ozair, Aaron Courville, and Yoshua Bengio.
\newblock Generative adversarial nets.
\newblock In {\em Proceedings of the 27th International Conference on Neural
  Information Processing Systems - Volume 2}, NIPS’14, page 2672–2680,
  Cambridge, MA, USA, 2014. MIT Press.

\bibitem{Larsen2016VAE_GAN}
Anders Boesen~Lindbo Larsen, Søren~Kaae Sønderby, Hugo Larochelle, and Ole
  Winther.
\newblock Autoencoding beyond pixels using a learned similarity metric, 2016.

\bibitem{Lin2020DualManifoldAdversarial}
Wei-An Lin, Chun~Pong Lau, Alexander Levine, Rama Chellappa, and Soheil Feizi.
\newblock Dual manifold adversarial robustness: Defense against lp and non-lp
  adversarial attacks, 2020.

\bibitem{Kruskal78}
J.B. Kruskal and M.~Wish.
\newblock {\em {Multidimensional Scaling}}.
\newblock Sage Publications, 1978.

\bibitem{Belkin03}
Mikhail Belkin and Partha Niyogi.
\newblock Laplacian eigenmaps for dimensionality reduction and data
  representation.
\newblock {\em Neural Computation}, 15:1373--1396, 2003.

\bibitem{Roweis00}
Sam~T. Roweis and Lawrence~K. Saul.
\newblock Nonlinear dimensionality reduction by locally linear embedding.
\newblock {\em SCIENCE}, 290:2323--2326, 2000.

\bibitem{Hinton03SNE}
Geoffrey Hinton and Sam Roweis.
\newblock Stochastic neighbor embedding.
\newblock {\em Advances in neural information processing systems}, 15:833--840,
  2003.

\bibitem{Maaten08T_SNE}
L.J.P. {van der Maaten} and G.E. Hinton.
\newblock Visualizing high-dimensional data using t-sne.
\newblock {\em Journal of Machine Learning Research}, 9(nov):2579--2605, 2008.
\newblock Pagination: 27.

\bibitem{ZhangHongyi17mix}
Hongyi Zhang, Moustapha Ciss{\'{e}}, Yann~N. Dauphin, and David Lopez{-}Paz.
\newblock mixup: Beyond empirical risk minimization.
\newblock {\em CoRR}, abs/1710.09412, 2017.

\bibitem{Verma19ICT}
Vikas Verma, Alex Lamb, Juho Kannala, Yoshua Bengio, and David Lopez-Paz.
\newblock Interpolation consistency training for semi-supervised learning,
  2019.

\bibitem{Verma2019ManifoldMixUp}
Vikas Verma, Alex Lamb, Christopher Beckham, Amir Najafi, Ioannis Mitliagkas,
  Aaron Courville, David Lopez-Paz, and Yoshua Bengio.
\newblock Manifold mixup: Better representations by interpolating hidden
  states, 2019.

\bibitem{YunSangdoo19}
Sangdoo, Dongyoon Han, Seong~Joon Oh, Sanghyuk Chun, Junsuk Choe, and Youngjoon
  Yoo.
\newblock Cutmix: Regularization strategy to train strong classifiers with
  localizable features.
\newblock {\em CoRR}, abs/1905.04899, 2019.

\bibitem{Berthelot19MixMatch}
David Berthelot, Nicholas Carlini, Ian Goodfellow, Nicolas Papernot, Avital
  Oliver, and Colin Raffel.
\newblock Mixmatch: A holistic approach to semi-supervised learning, 2019.

\bibitem{Yui19MixMatchImplementation}
Yui.
\newblock {\em Pytorch Implementation for Mix Match}, 2019.
\newblock imikushana@gmail.com.

\bibitem{Zagoruyko2016wideResNet}
Sergey Zagoruyko and Nikos Komodakis.
\newblock Wide residual networks, 2016.

\bibitem{Laine16}
Samuli Laine and Timo Aila.
\newblock Temporal ensembling for semi-supervised learning.
\newblock {\em CoRR}, abs/1610.02242, 2016.

\bibitem{Lee13PseudoLabel}
Dong hyun Lee.
\newblock Pseudo-label: The simple and efficient semi-supervised learning
  method for deep neural networks, 2013.

\bibitem{Tran19SESEMI}
Phi~Vu Tran.
\newblock Exploring self-supervised regularization for supervised and
  semi-supervised learning, 2019.

\bibitem{Tarvainen17MeanTeacher}
Antti Tarvainen and Harri Valpola.
\newblock Weight-averaged consistency targets improve semi-supervised deep
  learning results.
\newblock {\em CoRR}, abs/1703.01780, 2017.

\bibitem{Ke19DualStudent}
Zhanghan Ke, Daoye Wang, Qiong Yan, Jimmy Ren, and Rynson W.~H. Lau.
\newblock Dual student: Breaking the limits of the teacher in semi-supervised
  learning, 2019.

\bibitem{nair2019realmix}
Varun Nair, Javier~Fuentes Alonso, and Tony Beltramelli.
\newblock Realmix: Towards realistic semi-supervised deep learning algorithms,
  2019.

\bibitem{Wang19EnAET}
Xiao Wang, Daisuke Kihara, Jiebo Luo, and Guo-Jun Qi.
\newblock Enaet: Self-trained ensemble autoencoding transformations for
  semi-supervised learning, 2019.

\bibitem{berthelot2019remixmatch}
David Berthelot, Nicholas Carlini, Ekin~D. Cubuk, Alex Kurakin, Kihyuk Sohn,
  Han Zhang, and Colin Raffel.
\newblock Remixmatch: Semi-supervised learning with distribution alignment and
  augmentation anchoring, 2019.

\bibitem{Sohn20FixMatch}
Kihyuk Sohn, David Berthelot, Chun-Liang Li, Zizhao Zhang, Nicholas Carlini,
  Ekin~D. Cubuk, Alex Kurakin, Han Zhang, and Colin Raffel.
\newblock Fixmatch: Simplifying semi-supervised learning with consistency and
  confidence, 2020.

\bibitem{Tran2018DistGan}
Ngoc-Trung Tran, Tuan-Anh Bui, and Ngai-Man Cheung.
\newblock Dist-gan: An improved gan using distance constraints, 2018.

\end{thebibliography}

\clearpage
\section*{Appendix}
\label{sec:Appendix}

\subsection*{Projection for sum and positive} Proof, by using Lagrange multiplier, problem (\ref{equa:prox_operation_for_positif_and_constraint}) becomes:

\begin{equation*}
\begin{split}
\min_{\gamma,\mu \in \mathbb{R}^p,\lambda} \frac{1}{2} \sum_{i=1}^{p}\norm{\kappa_i-\gamma_i}_2^2 +&\lambda(\sum_{i=1}^{p}\gamma_i - c) + \sum_{i=1}^{p}\mu_i\gamma_i \\
\text{ subject to : } &\mu_1, \mu_2,.., \mu_p \leq 0 \\
\end{split}
\end{equation*}

We solve the following system of equations: \\
\vspace{0.2cm}
$\begin{cases}
  \gamma_i - \kappa_i+\lambda + \mu_i = 0 \\ 
  \sum_{i=1}^{p}\gamma_i = c\\
  \mu_i\gamma_i = 0 \\
  \mu_i \leq 0 \\
  \gamma_i \geq 0 \\
\end{cases}
\Leftrightarrow
\begin{cases}
  \lambda = \frac{1}{p}(\sum_{i=1}^p\kappa_i-\sum_{i=1}^p\mu_i-c) \\
  \gamma_i = \kappa_i-\frac{1}{p}(\sum_{i=1}^p\kappa_i-c) -\frac{p-1}{p}\mu_i + \frac{1}{p}\sum_{j\neq i}\mu_j  \\
  \sum_{i=1}^{p}\gamma_i = c\\
  \mu_i\gamma_i = 0 \\
  \mu_i \leq 0 \\
  \gamma_i \geq 0 \\
\end{cases}$\\
\vspace{0.2cm}

In the case that $ \kappa_i-\frac{1}{p}(\sum_{i=1}^p\kappa_i-c) < 0 $. From the second equation, we infer that $\mu_i \neq 0$ (because if $\mu_i = 0$ and $\mu_j \leq 0$ as in inequality 5, then $\gamma_i <0$, in contradiction to inequality 6). From $\mu_i \neq 0$, we infer that $\gamma_i = 0$ with equation 4. 

In the case that $ \kappa_i-\frac{1}{p}(\sum_{i=1}^p\kappa_i-c) = 0 $. From the second equation, first if $\mu_i = 0$ then $\gamma_i \leq 0$ since $\mu_j \leq 0$ as in equation 5. With inequality 6, we infer $\gamma_i = 0$. Second, if $\mu_i \neq 0$, then we infer that $\gamma_i = 0$ with equation 4.         

Let's $\mathcal{P} = \{i | \kappa_i-\frac{1}{p}(\sum_{i=1}^p\kappa_i-c) > 0 \}$ and $ \mathcal{N} = \{i | \kappa_i-\frac{1}{p}(\sum_{i=1}^p\kappa_i-c) \leq 0 \}$. We find exactly the same problem as before, but with only active index in the set $\mathcal{P}$.
\vspace{0.2cm}
\begin{center}
$\begin{cases}
  \gamma_i - \kappa_i+\lambda + \mu_i = 0, \forall i \in \mathcal{P} \\ 
  \sum_{i \in \mathcal{P}}\gamma_i = c\\
  \mu_i\gamma_i = 0, \forall i \in \mathcal{P} \\
  \mu_i \leq 0, \forall i \in \mathcal{P} \\
  \gamma_i \geq 0, \forall i \in \mathcal{P} \\
\end{cases}$
\end{center}
\vspace{0.2cm}

Then we repeat until the constraint satisfaction for $\gamma$. For a proof of convergence, as $\gamma$ has exactly $p$ elements $\gamma_i$, then each time we project to get a new active set $\mathcal{P}$, we reduce the number of active elements $\gamma_i$. As the number of active elements is something positive and it decreases, so it converges. Here is an implementation for multiple $\kappa$ ($\kappa \in \mathbb{R}^{M \times p}$) in \textit{pytorch}. 

\begin{lstlisting}
def prox_positive_and_sum_constraint(x,c):
    """ x is 2-dimensional array (M \times p) """ 
    
    n = x.size()[1]
    k = (c - torch.sum(x,dim=1))/float(n)
    x_0 = x + k[:,None]
    while len(torch.where(x_0 < 0)[0]) != 0:
        idx_negative = torch.where(x_0 < 0)        
        x_0[idx_negative] = 0.
        
        one = x_0 > 0
        n_0 = one.sum(dim=1)
        k_0 =(c - torch.sum(x_0,dim =1))/ n_0
        x_0 = x_0 + k_0[:,None] * one

    return x_0
\end{lstlisting}

\subsection*{Manifold attack for embedded representation}

Architecture of model $g()$ (\textit{Pytorch} style) :
\begin{itemize}
    \item[-] S curve data : \textit{Conv1d}[$1,4,2$] $\rightarrow$ \textit{ReLu} $\rightarrow$ \textit{Conv1d}[$4,4,2$] $\rightarrow$ \textit{ReLu}$\rightarrow$ \textit{Flatten} $\rightarrow$ \textit{Fc}[$4,2$].
    \item[-] Digit data : \textit{Conv2d}[$1,8,3$] $\rightarrow$ \textit{ReLu} $\rightarrow$ \textit{Conv2d}[$8,16,3$] $\rightarrow$ \textit{ReLu}$\rightarrow$ \textit{Flatten} $\rightarrow$ \textit{Fc}[$64,2$].
\end{itemize} 

Optimizer : Stochastic gradient descent, with learning rate $lr = 0.001$ and momentum = 0.9. Learning rate is reduce by $lr = lr^{0.5}$ after each 10 epochs. The number of epochs is 40.

\subsection*{Robustness to adversarial examples}

Hyper-parameters : optimizer = Stochastic gradient descent, number of epochs = 300, learning rate = 0.1, momentum = 0.9, learning rate is reduce by $lr = 0.1lr$ after each 75 epochs, batch size = 200, weight decay = 0.0001.

\subsection*{Semi-supervised manifold attack}

Hyper-parameters : optimizer = Adam, number of epochs = 1024, learning rate = 0.002, $\alpha = 0.75$, batch size labelled = batch size unlabelled = 64, T = 0.5 (in sharpening), $\lambda = 75 $ (linearly ramp up from 0), EMA = 0.999 , error validation after 1024 batchs. 

To reproduce an experiment, we define function seed\_ as:
\begin{lstlisting}
def seed_(p):
    """  for reproductive """
    torch.manual_seed(p)
    np.random.seed(p)
    random.seed(p)
    if torch.cuda.is_available():
        torch.cuda.manual_seed(p)
    torch.backends.cudnn.deterministic = True
    torch.backends.cudnn.benchmark = False 
            
    return 0
\end{lstlisting}

The four experiment 1,2,3,4 in \cref{tab:MixMatchAttack} are launched with respectively seed\_(0), seed\_(1), seed\_(2), seed\_(3). 

In Mix-up Adversarial, $n\_iters$ is fixed at 1. In dataset CIFAR-10, $\xi$ starts at 0.1 and decreases linearly to 0.01 after 1024 epochs. In dataset SVHN, $\xi$ starts at 0.1 and decreases linearly to 0.01 after 1024 epochs for seed\_(0) and seed\_(3); $\xi$ starts at 0.1 and decreases linearly to 0.001 after 1024 epochs for seed\_(2); $\xi$ starts at 0.01 and decreases linearly to 0.001 after 1024 epochs for seed\_(1)

\end{document}